\documentclass[11pt]{article}
\usepackage{agenticlearning-xelatex}
\usepackage{pifont}
\usepackage[bottom]{footmisc}

\renewcommand{\bm}[1]{\symbfit{#1}}
\usepackage{placeins}
\usepackage{etoolbox}
\AtBeginEnvironment{tabular}{\addfontfeatures{Numbers={Monospaced,Lining}}}
\newlength{\pmsdwidth}
\newcommand{\pmsd}[1]{\makebox[\pmsdwidth][l]{\scriptsize$\pm$#1}}
\newcommand{\plab}[3]{\llap{\makebox[\wd0][l]{\hspace{#1\wd0}\raisebox{#2\ht0}[0pt][0pt]{\small\textbf{(#3)}}}}}
\AtBeginDocument{\settowidth{\pmsdwidth}{\addfontfeatures{Numbers={Monospaced,Lining}}\scriptsize$\pm$00.0}}
\setlist[itemize]{leftmargin=*, itemsep=2pt, topsep=2pt, parsep=0pt}
\setlist[enumerate]{leftmargin=*, itemsep=2pt, topsep=2pt, parsep=0pt}
\graphicspath{{figures/}}
\crefname{proposition}{Proposition}{Propositions}
\Crefname{proposition}{Proposition}{Propositions}
\crefname{corollary}{Corollary}{Corollaries}
\Crefname{corollary}{Corollary}{Corollaries}
\newcommand{\E}{\mathbb{E}}
\newcommand{\R}{\mathbb{R}}
\newif\ifshowlimitations\showlimitationsfalse  %

\shorttitle{When JEPA Learns Latent Maps}
\shortauthor{Ren et al.}
\title{The Content--Geometry Tradeoff in World Models: \\When JEPA Learns Latent Maps}
\author{
  Mengye Ren$^{1}$, Ying Wang$^{*1}$, Christopher Hoang$^{*1}$, 
  Yilun Kuang$^{1}$, and Yann LeCun$^{12}$\\
  $^{1}$New York University \quad $^{2}$AMI Labs\\
  \texttt{\{mengye,yw3076,ch3451,yk2516,yann.lecun\}@nyu.edu}\\
  \url{https://agenticlearning.ai/content-geometry-tradeoff/}
}
\date{}

\usepackage{enumitem}
\begin{document}
\maketitle
\begingroup
\renewcommand{\thefootnote}{\fnsymbol{footnote}}
\makeatletter\renewcommand{\@makefntext}[1]{\noindent$^{\@thefnmark}$\hspace{0.3em}#1}\makeatother
\footnotetext[1]{Equal contribution.}
\endgroup

\begin{abstract}
\looseness=-1
A latent world model for planning needs two properties: latent distances that smoothly track progress toward a goal (\emph{geometry}), and task-relevant factors that remain decodable from the latent representation (\emph{content}). Common approaches prevent representation collapse with a stop-gradient (SG) target or a distributional regularizer, but we find that both often yield poor latent geometry.
In this paper, we introduce sliced variance regularization (SVReg), which imposes a variance floor on latent projections along random directions. 
SVReg learns a smooth map of the agent’s state. Its latent trajectories are smooth, and its latent distances track progress in navigation and reaching.
These geometric benefits, however, can come with a tradeoff in content. 
SVReg can discard slowly changing, hard-to-predict factors, such as the pose of a manipulated object, that SG variants tend to preserve. 
To address this tradeoff, we propose Slow SVReg, which favors slowly changing factors with a slowness term and a gated residual predictor. It keeps these factors and improves planning. This simple recipe has the highest planning success in the hard settings of five visual control tasks, and it learns latent maps from first-person mazes and city walks.
\end{abstract}

\section{Introduction}\label{sec:intro}
\begin{figure}[t]
\centering
\sbox0{\includegraphics[width=\linewidth]{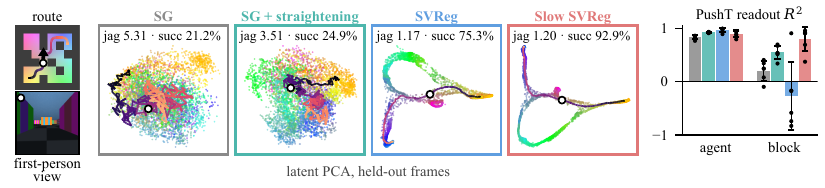}}\usebox0\plab{0}{0.8863}{a}\plab{0.7908}{0.8953}{b}
\caption{\textbf{SVReg learns a smooth map (geometry) but can lose the pushed block (content), and Slow SVReg keeps both.} (a)~One first-person route through the Medium maze in each latent, with the first-person view at the marked point. ``jag'' = the route's jaggedness (1 is smooth) and ``succ'' = planning success on hard goals. (b)~Held-out readout $R^2$ of the PushT agent and resting block, colored as in (a) (mean over 6 seeds).}
\label{fig:teaser}
\end{figure}

An autonomous agent inhabits a temporal world, observing a stream of high-dimensional sensory data from which it builds an internal model of how the world evolves. Learning latent dynamics from sequential observations is a natural path to such world models. An encoder compresses observations into a compact latent space, and a predictor captures temporal transitions, enabling planning and control without operating in the observation space. The Joint Embedding Predictive Architecture (JEPA)~\citep{lecun2022path} instantiates this idea. The JEPA objective alone, however, admits a trivial solution where the encoder maps everything to a constant. A popular fix today in modern self-supervised learning (SSL) is the \emph{stop-gradient} trick, which passes the prediction target through the encoder but blocks gradients on that branch. This is used by many JEPA implementations~\citep{assran2023self,bardes2024revisiting}, and the closely related image SSL methods~\citep{grill2020bootstrap,chen2021exploring}. Another alternative to stop-gradient is distributional regularizers~\citep{zbontar2021barlow,bardes2022vicreg,balestriero2025lejepa}. While the choice between the two is usually made for training stability, it also decides the structure of latent representations. A planner that searches in latent space~\citep{zhou2025dino,wang2026temporal,maes2026leworldmodel}
requires both \emph{geometry}, where latent distance smoothly tracks progress toward the goal, and \emph{content}, where task-relevant factors, such as an object to be manipulated, stay decodable.

We find that many common choices lead to poor geometry. With stop-gradient (SG), the target is held fixed during each update, so the encoder is trained only to predict its current latents. This one-sided objective does not ask consecutive latents to lie on a smooth path. In our experiments, a smooth route often appears jagged in the SG latent. 
Latent distance then becomes a poor guide toward far goals, and SG falls behind on hard maze goals. The one-sided objective is also fragile when data are scarce. In contrast to SG, common distributional regularizers fall short in a different way. VICReg and SIGReg were designed for i.i.d.\ multi-view batches and constrain the whole latent distribution. 

In contrast, a regularizer with fewer constraints yields a better latent map. In this paper, we introduce sliced variance regularization (SVReg), a one-sided floor on the spread of the latent along random directions, trained without SG. The target then moves with the encoder, so the whole latent is trained to be predictable from its past, and directions the predictor cannot explain, such as noise, are suppressed. In our experiments, a smooth route then stays smooth in the latent. The resulting latent is a smooth map of the agent's state. Its distances track progress in navigation and reaching, and in first-person mazes its leading principal components trace the agent's position.

However, the predictability pressure can erase task-relevant content. 
Factors are penalized for their unpredictable variance, so slow, hard-to-predict factors, such as manipulated objects, can be lost. 
In comparison, SG 
variants tend to
keep these factors, but at the cost of the jagged geometry described above. Together, the two failures form what we call the \emph{content--geometry tradeoff}. 

To achieve the best of both worlds, we propose Slow SVReg. It adds 1) a \emph{slowness} term, which encourages a small portion of the latent to change slowly across frames, and 2) a \emph{gated residual predictor}, which copies the current latent by default and predicts the change. Both favor slowly changing factors. 
Slow SVReg keeps these factors in manipulation and improves planning, especially in first-person mazes, where the agent's position must be carried across frames. Slow SVReg plans best on hard maze goals, Reacher, Cube and PushT. Moreover, a model trained on Wanderland city walk scans~\citep{liu2026wanderland} learns a latent map of the streets from first-person video and actions.

Our contributions are as follows. First, we show that SG and common distributional regularizers learn poor latent geometry for planning, while SVReg learns smooth latent maps in navigation and reaching. Second, we identify the content--geometry tradeoff, analyze its direction in the linear encoder objective, show how modes compete for latent capacity in controlled experiments on latent linear dynamics, and find the same pattern in visual control. Finally, we introduce the slowness regularizer and the gated residual predictor, which favor slowly changing factors and keep content more reliably. The resulting model performs strongly across navigation, mapping, reaching, and manipulation. 

\section{Analysis}\label{sec:theory}
This section explains where the content--geometry tradeoff comes from and motivates our main method in \Cref{sec:method}, through the lens of linear latent dynamics. Without stop-gradient, the encoder is penalized for unpredictable variance (\Cref{sec:dichotomy}). This suppresses noise, but it also shrinks hard-to-predict factors, which can then be dropped when the latent has too few dimensions to hold every factor (\Cref{sec:content}). With stop-gradient, finite data can lead the encoder to read noise while its latent statistics look healthy (\Cref{sec:collapse}). 

We analyze $T$ observations with a linear encoder $\Phi(\bm{o})=W\bm{o}$ and an action-conditioned linear predictor $\hat{\bm{z}}_{t+1}=A\bm{z}_t+B\bm{a}_t$, where $\bm{z}_t=\Phi(\bm{o}_t)$. The JEPA loss~\citep{lecun2022path} is
\[
  \mathcal{L} = \frac{1}{T-1}\sum_{t=0}^{T-2}
  \bigl\|A\bm{z}_t+B\bm{a}_t-\bm{z}_{t+1}\bigr\|^2.
\]
This loss admits a trivial collapse in which $\Phi$ maps every observation to the same constant ($\Phi\equiv0$ for the linear encoder). The remedies are a stop-gradient on the target~\citep{assran2023self,bardes2024revisiting}, or a regularizer on the latent distribution~\citep{zbontar2021barlow,bardes2022vicreg,balestriero2025lejepa,kuang2026lpwm}.

\subsection{Two JEPA Objectives}\label{sec:dichotomy}
For mean-centered observations, define the lag-0 ($\hat\Sigma_{00}$), next-step ($\hat\Sigma_{11}$), and lag-1 ($\hat\Sigma_{01}$) sample covariances
\[
  \hat\Sigma_{00}=\frac{\sum_t\bm{o}_t\bm{o}_t^\top}{T-1},\quad
  \hat\Sigma_{11}=\frac{\sum_t\bm{o}_{t+1}\bm{o}_{t+1}^\top}{T-1},\quad
  \hat\Sigma_{01}=\frac{\sum_t\bm{o}_t\bm{o}_{t+1}^\top}{T-1},
\]
with sums over $t=0,\ldots,T-2$, latent statistics $\Sigma_z=W\hat\Sigma_{00}W^\top$ and $C_z=W\hat\Sigma_{01}W^\top$, and $\Sigma_z$ invertible (no latent direction is constant).

\begin{theorem}[Encoder objective after eliminating the predictor]\label{thm:dichotomy}
For a linear encoder $\Phi(o)=Wo$ and a predictor without the action term, substituting the optimal linear predictor gives:
\begin{enumerate}
  \item \textbf{Without stop-gradient}, $A^\star = C_z^\top\Sigma_z^{-1}$ and
  \[
    \mathcal{L}^\star_{\mathrm{no\text{-}sg}}(W) \;=\; \mathrm{tr}\!\bigl(W\hat\Sigma_{11}W^\top\bigr) \;-\; \mathrm{tr}\!\bigl(C_z^\top \Sigma_z^{-1} C_z\bigr).
  \]
  \item \textbf{With stop-gradient and a frozen target $W_{\mathrm{tgt}}$}, writing $\hat\Sigma_{zz'} = W\hat\Sigma_{01}W_{\mathrm{tgt}}^\top$, $A^\star = \hat\Sigma_{zz'}^\top\Sigma_z^{-1}$ and
  \[
    \mathcal{L}^\star_{\mathrm{sg}}(W;\,W_{\mathrm{tgt}}) \;=\; \mathrm{tr}\!\bigl(W_{\mathrm{tgt}}\hat\Sigma_{11}W_{\mathrm{tgt}}^\top\bigr) \;-\; \mathrm{tr}\!\bigl(\hat\Sigma_{zz'}^\top\Sigma_z^{-1}\hat\Sigma_{zz'}\bigr).
  \]
\end{enumerate}
\end{theorem}

The proof is in \Cref{app:proofs}, and \Cref{app:actions} extends the theorem to the action-conditioned predictor. Without stop-gradient, the loss also penalizes unexplained target variance. Removing unpredictable directions, including observation noise, therefore reduces the loss and can give a smoothly moving state a smooth latent path. With stop-gradient (SG), target variance is constant during an update, so noise directions are unpenalized and perfectly predictable rotations allow curved paths even at zero prediction error.

\subsection{Finite Data}\label{sec:collapse}
Finite-sample noise has a nonzero lag-1 cross-covariance, of order $\sigma^2\sqrt{d/T}$ for observation dimension $d\ll T$ and noise level $\sigma$. When this spurious lag structure exceeds the signal's, the fixed-target objective can select it, and the encoder keeps healthy latent variance while reading noise instead of state. We call this failure \emph{signal collapse}. In-sample statistics cannot detect it, since the selected noise directions mimic signal lag correlations and a variance regularizer rescales them to meet its floor. A held-out state probe does detect it (Proposition~\ref{prop:stable} in \Cref{app:prop-discussion}).

\subsection{Content}\label{sec:content}
While good geometry captures latent dynamics, task-relevant content such as the state of a manipulated object can still be lost. Consider one scalar component of the environment state that evolves as $s_{t+1}=\varphi s_t+\kappa a_t+\epsilon_t$. Its persistence $\varphi$ measures how much of it carries over to the next step, so a mode with $\varphi$ near $1$ changes slowly. Its one-step residual $\delta$ is the share of its variance that is not predicted by the current state and action. Without stop-gradient, a less predictable mode adds more prediction loss, and SVReg shrinks its latent coordinate. 

The variance floor keeps the mode's scale away from zero, so shrinkage alone keeps the mode readable (\Cref{app:content}). Content is lost when modes compete for too few latent dimensions. We analyze this competition in the linear case, where two modes share one coordinate (Proposition~\ref{prop:content} in \Cref{app:content}). For persistent modes, with or without stop-gradient, a new mode enters the latent more easily when it is more persistent than the mode already there. Without stop-gradient, it also enters more easily when the action explains more of its variance, and the cost of one coordinate carrying two dynamics can protect the incumbent, so whichever factor fills the latent first can keep it. A manipulated object rests most of the time and moves only through contact, so the action explains little of its variance, and the latent can trade it for an action-driven factor such as the arm. SVReg loses the cube's height early in training, as this competition predicts (\Cref{fig:cube_height}b). In \Cref{app:toys} we also test the competition of capacity with a linear experiment (\Cref{fig:toy_content}).

\section{Learning Geometry and Content}\label{sec:method}
\begin{figure}[t]
\centering
\sbox0{\includegraphics[width=\linewidth]{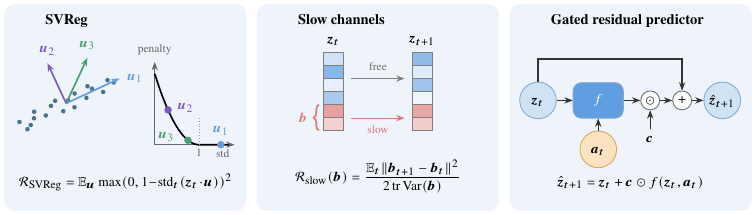}}\usebox0\plab{0.0213}{0.8841}{a}\plab{0.3556}{0.8841}{b}\plab{0.6900}{0.8841}{c}
\caption{The three components of Slow SVReg. (a)~SVReg penalizes random directions $\bm{u}$ whose latent standard deviation $\mathrm{std}_t(\bm{z}_t\cdot\bm{u})$ falls below $1$ ($\bm{u}_2$, $\bm{u}_3$). 
(b)~The slow channels $\bm{b}_t$, a fixed set of latent dimensions (red), are penalized by the variance-normalized slowness term (\Cref{eq:slow}). The other dimensions change freely. (c)~The gated residual predictor copies $\bm{z}_t$ and adds the change predicted by $f$ from the latents and the action $\bm{a}_t$, scaled by the learned gate $\bm{c}$ (\Cref{eq:gate}).}
\label{fig:method}
\end{figure}

In this section, we introduce Slow SVReg. It trains without stop-gradient with sliced variance regularization (SVReg) for geometry, and adds \emph{slow channels} and a \emph{gated residual predictor} to favor slowly changing factors (\Cref{fig:method}). We also define measures of content and geometry. The baselines are described in \Cref{sec:visual:setup}.

\paragraph{Sliced variance regularization.} VICReg and SIGReg (\Cref{sec:visual:setup}) were designed for i.i.d.\ multi-view batches and constrain the whole latent distribution, toward decorrelated coordinates or an isotropic Gaussian. World model data are temporally correlated trajectories whose states need not follow such a distribution, even with ample data, which motivates a less restrictive regularizer. Sliced variance regularization (SVReg) puts a one-sided floor on the standard deviation of the latent along random directions:
\begin{equation}\label{eq:floor-sl}
  \mathcal{R}_{\mathrm{SVReg}}(z) = \E_{\bm{u}}\!\left[\max\!\left(0,\,1-\mathrm{std}_t(\bm{z}_t \cdot \bm{u})\right)^2\right],
  \qquad \bm{u}\sim\operatorname{Unif}(\mathbb{S}^{r-1}),
\end{equation}
where $\bm{z}_t \cdot \bm{u}$ projects each latent onto the direction $\bm{u}$ and $\mathrm{std}_t$ is taken over the estimation sample. The squared hinge is active only when the standard deviation along the direction falls below $1$. Restricting $\bm{u}$ to the coordinate axes gives the element-wise floor, a squared form of the variance term of VICReg~\citep{bardes2022vicreg}. A latent can satisfy the element-wise floor while two coordinates copy each other, leaving no variance along their difference. SVReg also checks such mixed directions, without requiring the coordinates to be decorrelated.
We add SVReg with weight $\lambda$ to the JEPA prediction loss without stop-gradient, just like VICReg and SIGReg.

\paragraph{Slow channels.}
SVReg alone can lose content that rarely changes and moves only through contact, such as a resting cube. We therefore train a small fixed set of latent channels to change slowly between frames. For the slow channels $\bm{b}_t$, we add the slowness term with weight $\beta$,
\begin{equation}\label{eq:slow}
  \mathcal{R}_{\mathrm{slow}}(\bm{b})
  = \frac{\E_t\|\bm{b}_{t+1}-\bm{b}_t\|^2}
  {2\,\mathrm{tr}\,\mathrm{Var}(\bm{b})}.
\end{equation}
This is a variance-normalized form of the slowness objective of slow feature analysis~\citep{wiskott2002slow}, so shrinking the slow channels cannot reduce the term, while SVReg covers the whole latent. It resembles the invariance term of VICReg applied to consecutive frames, but acts on a subset of channels and is normalized by their variance. Prior world models apply slowness along trajectories to the whole latent, in JEPA objectives such as PLDM~\citep{sobal2025learning,hauri2026learning} and as a regularizer that keeps consecutive latents close~\citep{xia2026cloning}. The closest split designs constrain a fixed part of a representation~\citep{garrido2023self,denton2017unsupervised}.

\paragraph{Gated residual predictor.}
The gated residual predictor is a second way to favor slow factors. It copies the current latent by default and learns the change, so a factor that does not change between frames costs nothing to predict. The linear dynamics of \Cref{sec:content} already have this form, $s_{t+1} = s_t + \bigl((\varphi-1)s_t + \kappa a_t\bigr) + \epsilon_t$. For a persistent mode, $\varphi$ near $1$ makes the change without action small, as for a resting object or the agent's position between moves. We use
\begin{equation}\label{eq:gate}
  \hat{\bm{z}}_{t+1}
  = \bm{z}_t + \bm{c}\odot f(\bm{z}_{\le t},\bm{a}_t),
\end{equation}
where $\bm{a}_t$ is the action and $\bm{c}$ is a learned per-dimension gate initialized at $c_0 = 0.1$. This has the same form as LayerScale~\citep{touvron2021going}, which scales residual branches across depth, whereas this gate scales the predicted change over time.

The identity path makes slowly changing factors cheap to predict, while the slow channels add direct pressure toward slow change. Together, they are designed to favor slow factors without SG, echoing SG's local preference among competing modes. We call this model Slow SVReg.

\begin{figure}[!tb]
\centering
\begin{minipage}{0.92\linewidth}
\sbox0{\includegraphics[width=\linewidth]{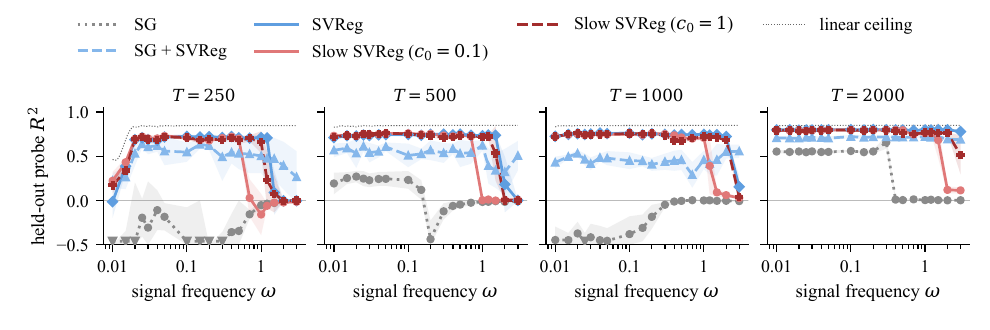}}\usebox0\plab{0}{0.9075}{a}\\[0.2em]
\sbox0{\includegraphics[width=\linewidth]{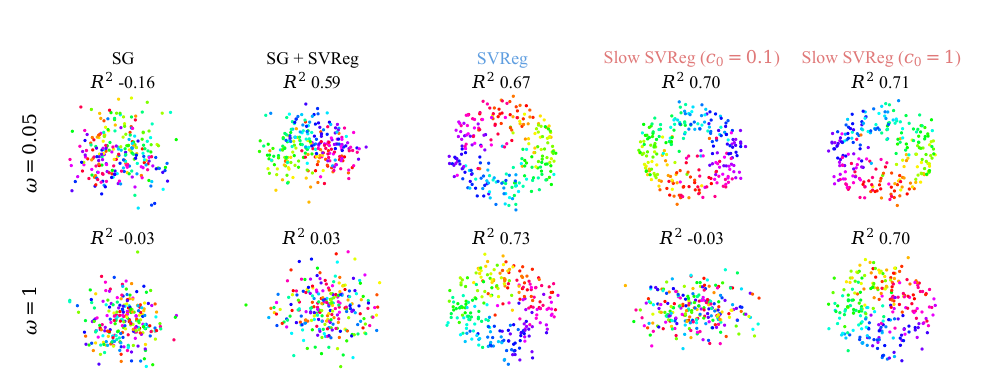}}\usebox0\plab{0}{0.8352}{b}
\end{minipage}
\caption{\textbf{Linear dynamics: SVReg recovers slow and fast signals and keeps the circle structure.} (a)~\label{fig:toy_geometry}Held-out probe $R^2$ of the circle state against signal frequency at various trajectory length $T$ (mean over 8 seeds). (b)~\label{fig:toy_latent_map}Top two principal components of held-out latents ($T{=}250$, one seed), colored by circle phase. Protocol in \Cref{app:toys}.}
\end{figure}

\paragraph{Measuring content and geometry.}\label{sec:measures} We measure both properties on held-out data, separately from planning success. \emph{Content} is the held-out $R^2$ of ridge readouts of state factors (agent position, joint angles, cube position and height, block pose), fit and scored on separate episodes,
 and the area under the ROC curve for binary factors such as gripper--cube contact ($0.5$ is chance). We measure two aspects of \emph{geometry}, the ordering of latent distances and the smoothness of latent paths. The \emph{distance rank correlation} $\rho$ is the Spearman correlation between latent distance $\|\bm{z}-\bm{z}'\|$ and true task distance (shortest path in PointMaze, joint space in Reacher). In manipulation, \emph{cost-to-goal $\rho$} correlates the latent L2 distance between two held-out frames (the planner's cost) with their true state distance (\Cref{app:setup}). \emph{Jaggedness} encodes a scripted smooth motion, such as a maze route or a full Reacher shoulder turn, and compares path lengths before and after a short moving average $\bar{\bm{z}}_t$,
\begin{equation}\label{eq:jag}
  J = \frac{\sum_t \|\bm{z}_{t+1}-\bm{z}_t\|}{\sum_t \|\bar{\bm{z}}_{t+1}-\bar{\bm{z}}_t\|},
\end{equation}
so smooth paths give $J \approx 1$. The \emph{participation ratio} (PR, \emph{eff.\ dim} in figures and tables) counts the directions the latent effectively uses. With $\mu_1,\dots,\mu_r$ the eigenvalues of the held-out latent covariance,
\begin{equation}\label{eq:pr}
  \mathrm{PR}=\frac{\bigl(\sum_i\mu_i\bigr)^2}{\sum_i\mu_i^2}\in[1,r].
\end{equation}

\section{Noisy Linear Dynamics}\label{sec:toys}

We first test the geometry prediction of \Cref{sec:theory} in a linear world whose map is known. A two-dimensional state rotates at frequency $\omega$ and is observed through a random orthonormal map $U\in\R^{32\times2}$ with Gaussian noise,
\[
  \bm{s}_{t+1} = R_\omega\,\bm{s}_t, \qquad \bm{o}_t = U\bm{s}_t + \sigma\bm{n}_t, \quad \bm{n}_t\sim\mathcal{N}(0, I),
\]
where $R_\omega$ is the rotation by angle $\omega$ and $\sigma=0.3$ (protocol in \Cref{app:toys}). The circle is an easy task, since a PCA of the raw observations can recover the latent easily. The experiment is therefore a unit test that asks whether a JEPA objective can also find a latent linear map.

Interestingly, SG fails most at fast frequencies and short trajectories, as shown in \Cref{fig:toy_geometry}. By comparison, SVReg stays near the signal-recovery ceiling up to $\omega\approx1$, and SG + SVReg lies between them. Slow SVReg matches SVReg on slow signals. With its default small gate ($c_0{=}0.1$) it loses the fastest signals (above $\omega=0.5$ at $T{=}250$), and a gate of $c_0{=}1$ keeps them up to $\omega\approx1$. 
In \Cref{fig:toy_latent_map}, we show the different learned latent maps. Each encoder maps a loop, as an agent walking a closed route would. Without SG the map is a ring ordered by phase, while SG encodes scattered points without circular geometry.

The dependence on trajectory length follows the finite-data argument of \Cref{sec:collapse}. The spurious lag structure of the noise shrinks as $\sigma^2\sqrt{d/T}$, so a frozen-target encoder can select noise directions on short trajectories and needs longer ones to recover the circle. Without stop-gradient, SVReg penalizes unexplained target variance and so removes the noise directions at every length. The analysis does not predict the dependence on frequency, which we report as an empirical finding.

A second linear experiment tests content under limited capacity (\Cref{app:toys}). An action-driven arm and an object that moves only when the arm touches it compete for a latent smaller than their joint state. SVReg keeps the arm with the best geometry but drops the object in about half the seeds (\Cref{fig:toy_content}). SG keeps the object at the cost of arm geometry, and SG + SVReg lies between them. 

\section{Visual Control}\label{sec:visual}

In this section, we test whether geometry and content decide planning success in visual goal-reaching with an action-conditioned JEPA. In navigation and reaching the agent moves only itself, and in manipulation it also moves an object.

\subsection{Setup}\label{sec:visual:setup}

\textbf{Environments.} We evaluate PointMaze U-maze and Medium~\citep{fu2020d4rl}, Reacher-Hard~\citep{tassa2018deepmind}, OGBench-Cube~\citep{park2025ogbench} and PushT~\citep{chi2025diffusion}. Top-down PointMaze and Reacher train on one uniform-random exploration trajectory, first-person PointMaze on one trajectory of smooth random driving, Cube on 2{,}000 scripted-expert episodes and PushT on DINO-WM's offline dataset~\citep{zhou2025dino}. 

\textbf{Models and baselines.} All models trained from scratch share a task-specific ResNet encoder and an action-conditioned transformer predictor (architectural details in \Cref{app:arch}). VICReg~\citep{bardes2022vicreg} combines element-wise variance and covariance penalties. As in other JEPA world models, it is paired with action-conditioned next-latent prediction, so on temporal data it has no term that keeps consecutive latents close. SIGReg, the sketched isotropic Gaussian regularizer~\citep{balestriero2025lejepa}, matches Gaussianity along random slices. 
As in LeWM~\citep{maes2026leworldmodel}, we apply SIGReg after a BatchNorm MLP projector on the encoder output and call this SIGReg-BN. The projector is the only difference from plain SIGReg, and the other objectives act on the encoder output directly. LeWM computes the SIGReg statistics per frame, while LeJEPA pools them over the batch. To give SIGReg-BN its best chance, we ran both forms on every task and use the one that plans better (\Cref{app:setup}). Per-frame statistics plan better on Cube, Reacher and the first-person mazes, and statistics pooled over the batch and time on the top-down mazes and PushT.
Two hybrids test whether explicit regularization yields geometry with SG. \emph{SG + SVReg} adds SVReg (\Cref{eq:floor-sl}), and \emph{SG + straightening} adds the temporal-straightening curvature penalty~\citep{wang2026temporal}
\begin{equation}\label{eq:curv}
  \mathcal{R}_{\mathrm{curv}}(z) = \frac{1}{T-2}\sum_{t=0}^{T-3}\Bigl(1-\cos\bigl(\bm{z}_{t+1}-\bm{z}_t,\;\bm{z}_{t+2}-\bm{z}_{t+1}\bigr)\Bigr)
\end{equation}
over a training window of length $T$, which is $0$ along a straight latent path.

\textbf{Planning and selection.} Cross-entropy method (CEM) model-predictive control minimizes latent distance from predicted states to the goal image and replans after a few steps, as in DINO-WM~\citep{zhou2025dino}. In the mazes, easy goals lie 25 steps from the start with no filter on their distance, and hard goals lie 50 steps away and at least 3 maze units apart. Cube is scored on all goals and on the goals that move the cube by at least 4\,cm ($\geq$4\,cm, about 60\% of goals). For each method, task and goal set we select one encoder learning rate and regularization weight, the setting with the highest mean success on a separate planning seed over all training seeds. We report that setting on three other planning seeds (\Cref{app:setup}). Unless stated otherwise, tables report the mean and standard deviation over six training seeds at the selected setting. Bold marks the best value and underline the second best.

\subsection{Main Results}\label{sec:visual:main}
\begin{table}[!t]\centering\small\setlength{\tabcolsep}{4pt}
\caption{\textbf{Planning success (\%) in five visual control tasks.} Mean $\pm$ s.d.\ over six training seeds.}

\label{tab:main}  %
\resizebox{\linewidth}{!}{%
\begin{tabular}{l*{8}{c}}
\toprule
 & \multicolumn{5}{c}{Agent only: navigation and reaching} & \multicolumn{3}{c}{Object manipulation} \\
\cmidrule(lr){2-6} \cmidrule(lr){7-9}
 & \multicolumn{2}{c}{U-maze (5k)} & \multicolumn{2}{c}{Medium (20k)} & Reacher (5k) & \multicolumn{2}{c}{Cube (80k)} & PushT (470k) \\
{\scriptsize steps $\times$ batch} & \multicolumn{2}{c}{\scriptsize 10k $\times$ 32} & \multicolumn{2}{c}{\scriptsize 20k $\times$ 32} & {\scriptsize 50k $\times$ 32} & \multicolumn{2}{c}{\scriptsize 50k $\times$ 64} & {\scriptsize 50k $\times$ 32} \\
\cmidrule(lr){2-3} \cmidrule(lr){4-5} \cmidrule(lr){6-6} \cmidrule(lr){7-8} \cmidrule(lr){9-9}
Method & easy & hard & easy & hard & & all & $\geq$4\,cm & \\
\midrule
\multicolumn{9}{l}{\emph{Stop-gradient}} \\
SG                            & 85.1\pmsd{17.0} & 65.0\pmsd{18.7} & 96.6\pmsd{3.1} & 71.6\pmsd{8.7} & 86.5\pmsd{1.6} & 80.4\pmsd{2.1} & 67.2\pmsd{3.6} & 44.0\pmsd{19.3} \\
SG + straightening            & 86.3\pmsd{5.5} & 88.0\pmsd{3.1} & 96.9\pmsd{1.6} & 72.8\pmsd{5.6} & 86.6\pmsd{2.9} & 85.4\pmsd{1.0} & 75.6\pmsd{1.6} & 62.1\pmsd{7.7} \\
SG + SVReg                    & 96.3\pmsd{1.6} & 95.8\pmsd{1.1} & 96.0\pmsd{1.5} & 62.7\pmsd{6.2} & 96.3\pmsd{1.6} & 77.2\pmsd{1.3} & 61.7\pmsd{2.3} & \underline{76.9}\pmsd{4.3} \\
\midrule
\multicolumn{9}{l}{\emph{No stop-gradient}} \\
VICReg                        & \textbf{99.9}\pmsd{\textbf{0.3}} & 94.3\pmsd{2.3} & 96.4\pmsd{0.8} & 77.2\pmsd{2.3} & 93.1\pmsd{1.7} & 69.6\pmsd{5.8} & 49.8\pmsd{9.3} & 55.0\pmsd{3.6} \\
SIGReg-BN                     & 92.4\pmsd{7.9} & 82.2\pmsd{13.7} & 97.3\pmsd{1.0} & 78.2\pmsd{3.8} & 95.8\pmsd{1.3} & \underline{87.4}\pmsd{0.8} & \underline{79.0}\pmsd{1.3} & 60.4\pmsd{2.6} \\
\rowblue SVReg (ours)                  & \underline{99.7}\pmsd{0.4} & \underline{96.4}\pmsd{1.3} & \textbf{98.0}\pmsd{\textbf{1.3}} & \underline{81.3}\pmsd{2.3} & \underline{98.7}\pmsd{0.9} & 76.9\pmsd{8.8} & 62.9\pmsd{13.1} & 75.0\pmsd{8.6} \\
\rowblue Slow SVReg (ours)             & \underline{99.7}\pmsd{0.4} & \textbf{98.7}\pmsd{\textbf{0.7}} & \underline{97.9}\pmsd{0.8} & \textbf{84.7}\pmsd{\textbf{2.7}} & \textbf{99.6}\pmsd{\textbf{0.4}} & \textbf{89.1}\pmsd{\textbf{1.6}} & \textbf{81.9}\pmsd{\textbf{2.5}} & \textbf{87.3}\pmsd{\textbf{3.8}} \\
\bottomrule
\end{tabular}}
\end{table}

\begin{table}[!t]\centering\small\setlength{\tabcolsep}{5pt}
\caption{\textbf{Planning success (\%) in first-person maze navigation.} Mean $\pm$ s.d.\ over six training seeds. Setup in \Cref{app:setup}.}
\label{tab:ego}
\begin{tabular}{lcccc}
\toprule
 & \multicolumn{2}{c}{U-maze (5k frames)} & \multicolumn{2}{c}{Medium (20k frames)} \\
\cmidrule(lr){2-3} \cmidrule(lr){4-5}
Method & easy & hard & easy & hard \\
\midrule
SG & 97.2\pmsd{1.1} & 42.0\pmsd{6.8} & 98.4\pmsd{0.9} & 21.2\pmsd{6.4} \\
SG + straightening & 96.6\pmsd{1.5} & 37.7\pmsd{16.0} & 97.2\pmsd{1.9} & 24.9\pmsd{8.8} \\
SG + SVReg & 97.6\pmsd{1.2} & 38.3\pmsd{12.5} & 97.2\pmsd{4.1} & 46.0\pmsd{16.6} \\
VICReg & 94.2\pmsd{1.2} & 40.3\pmsd{4.4} & 95.0\pmsd{1.0} & 29.6\pmsd{7.3} \\
SIGReg-BN & \underline{99.0}\pmsd{0.8} & 71.0\pmsd{6.1} & \underline{98.7}\pmsd{0.8} & 42.8\pmsd{5.2} \\
\rowblue SVReg & 98.2\pmsd{0.8} & \underline{90.4}\pmsd{1.6} & 97.0\pmsd{0.7} & \underline{75.3}\pmsd{5.5} \\
\rowblue Slow SVReg & \textbf{99.7}\pmsd{0.6} & \textbf{95.6}\pmsd{1.4} & \textbf{99.4}\pmsd{0.5} & \textbf{92.9}\pmsd{3.2} \\
\bottomrule
\end{tabular}
\end{table}

Slow SVReg plans best on hard maze goals from both views, on Reacher, on both Cube protocols and on PushT (\Cref{tab:main,tab:ego}). In the mazes, easy goals nearly saturate for the strongest methods, while hard goals lie 50 steps away along long routes and test whether latent distance tracks progress toward the goal. SVReg alone shows where the two properties matter. It plans second only to Slow SVReg on hard goals in navigation and reaching, where only the agent moves, but it falls behind SIGReg-BN on Cube and SG + SVReg on PushT, with widely varying success. We trace its strength in navigation to geometry (\Cref{sec:visual:geometry}) and its weakness with objects to lost content (\Cref{sec:visual:content}).

In \Cref{tab:ego}, we report planning in first-person view mazes, and the gaps widen. Here, the agent sees only the view ahead, so its position must be inferred from the views and actions. On hard goals SG drops sharply and the other baselines stay well below SVReg, while Slow SVReg nearly saturates on both mazes. Among the agent-only tasks, its gain over SVReg is largest here and small in the top-down mazes and Reacher, where each frame shows the agent.

SG + SVReg and SG + straightening add a geometry term to SG. They plan better than SG on hard U-maze goals and on PushT, but trail SVReg on hard Medium goals in both views and on first-person U-maze.

\subsection{Geometry}\label{sec:visual:geometry}

\begin{figure}[t]
\centering
\includegraphics[width=0.92\linewidth]{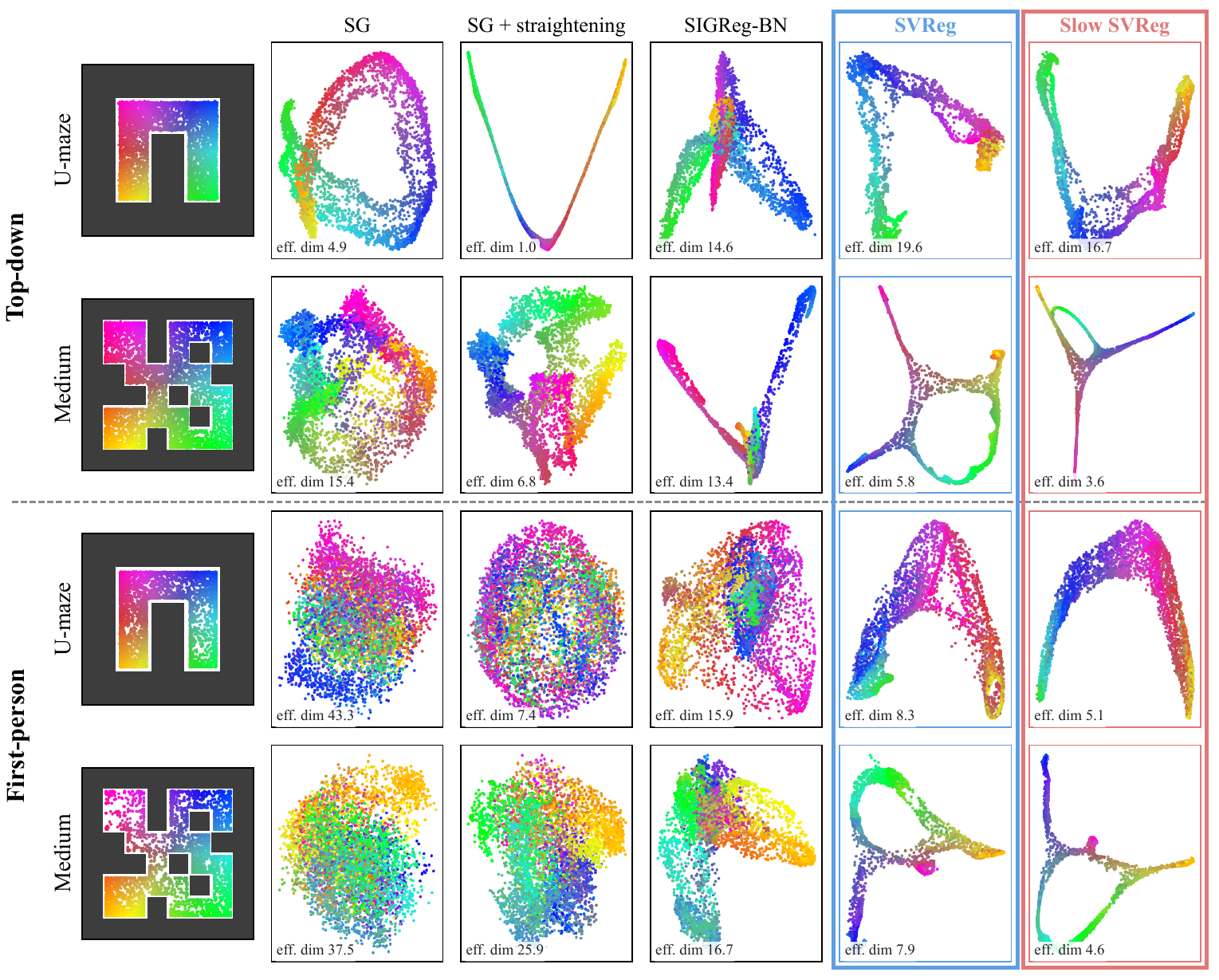}
\caption{\textbf{Without stop-gradient, the latent traces the maze from top-down and first-person views.} Held-out PCA latents colored by position, with SVReg boxed in blue and Slow SVReg in red. First-person panels are shown for the methods in \Cref{tab:ego}. Setups in \Cref{app:setup}.}
\label{fig:maze_pca}
\end{figure}

\Cref{fig:maze_pca} plots the top two principal components of held-out latents in both mazes, colored by the agent's position. Without SG, SVReg and Slow SVReg show the maze layout in the PCA plot. Nearby positions are close in the representation space, and the corridors of Medium appear as separate branches. The layout emerges from first-person views as well, where the agent never sees its true position. 
SG keeps only a coarse ordering by position from above, and with first-person views its latents, like those of SG + straightening, mix positions from across the maze. 
The SVReg paths are smooth as well. A smooth shortest route stays smooth in most SVReg latents and zigzags under SG (\Cref{app:pca}), and a full turn of the Reacher shoulder traces a smooth closed loop under SVReg and a jagged one under SG.

A latent map can be right globally and still be hard to plan with. 
The planner reaches a far goal by following latent distance over many short replanning steps, so it depends on the local shape of the latent and not only on its global layout. 
We therefore compare planning with two measures of geometry: the jaggedness of latent paths, a local measure, and the rank correlation between latent distance and maze distance over all frame pairs, a global measure. Jaggedness tracks planning closely. On U-maze, Medium and Reacher, the three methods with the most jagged paths are also the three weakest planners (\Cref{tab:maze_geometry,tab:reacher_geometry}). 
The rank correlation does not track planning. SG + SVReg plans hard U-maze goals well although its latent distance barely follows the maze. Planning success follows path smoothness more closely than global distance ranking, which fits a planner that compares latent distances over a few steps.

\subsection{Content}\label{sec:visual:content}

\begin{figure}[t]
\centering
\begin{minipage}[t]{0.495\linewidth}
\sbox0{\includegraphics[width=\linewidth]{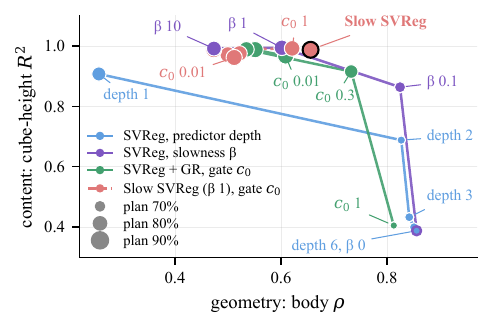}}\usebox0\plab{0}{0.8995}{a}
\end{minipage}\hfill
\begin{minipage}[t]{0.495\linewidth}
\sbox0{\includegraphics[width=\linewidth]{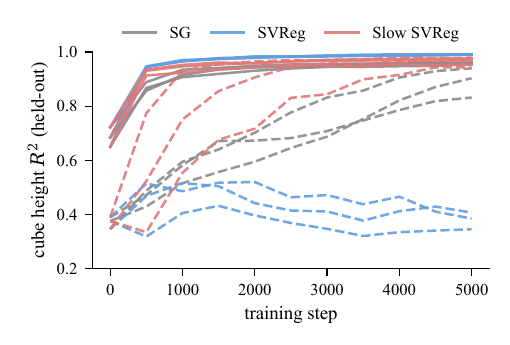}}\usebox0\plab{0}{0.8892}{b}
\end{minipage}
\caption{\textbf{Cube content against geometry, and cube-height acquisition.} (a)~Each curve varies predictor depth, slowness weight $\beta$, or the fixed gate $c_0$ of the gated residual predictor (GR), the last without and with slow channels. Gate curves use runs in which GR alone trains well. $\rho$: rank correlation between latent and effector distances. Circle area: planning success. Black edge: Slow SVReg. (b)~Cube-height readout early in training, dashed for seeds with a weak initial readout (\Cref{app:setup}).}
\label{fig:cube_height}
\end{figure}

Geometry alone does not explain the object tasks. On Cube, SIGReg-BN and SG + straightening plan second and third in \Cref{tab:main} although their latent distances follow goal distance least well, while SVReg ranks goal distance well and still plans poorly. Here the planners differ in content, and in particular in the cube rather than the arm (\Cref{tab:cube_geometry}). VICReg, SVReg and SG + SVReg, the weakest Cube planners, detect gripper--cube contact least reliably, and SVReg can lose the cube's height (\Cref{fig:cube_height}b).

PushT shows the same effect more sharply. A resting block changes only when the agent pushes it, so it is hard to predict from the agent's motion. VICReg, SIGReg-BN and SVReg all lose it, and their held-out block readouts fall below predicting the mean (\Cref{tab:pusht_geometry}). The two best PushT planners in \Cref{tab:main}, Slow SVReg and SG + SVReg, both keep the block.

Keeping this content usually costs geometry. Along single knobs on Cube, such as predictor depth or the slowness weight, a higher cube-height readout comes with a lower body $\rho$ (\Cref{fig:cube_height}a). Slow SVReg reaches the high-content end of this tradeoff and gives up little geometry. It reliably keeps cube height and contact, keeps the PushT block in most runs (\Cref{fig:teaser}b), and its cost-to-goal $\rho$ stays close to SVReg's on both tasks. Two-stage, an SVReg head trained on a frozen SG encoder, keeps content but plans below Slow SVReg on all tasks (\Cref{app:twostage}).

\subsection{Content Is Decided Early}\label{sec:visual:early}

The random initial encoder decides which SVReg runs lose cube height. For each training seed, the three objectives start from the same encoder, and SVReg fails exactly from the initial encoders that read the least height. From these starts, SG keeps acquiring height while SVReg does not (\Cref{fig:cube_height}b). In shorter reruns, Slow SVReg acquires height from all starts, more slowly from the weakest ones (\Cref{fig:cube_height}b).

Once acquired, height survives the removal of SG. In \Cref{app:switch}, switching from SG to SVReg at 20k steps keeps cube height and plans better than SG. SVReg's content failure therefore lies in acquisition.

\subsection{Ablation}\label{sec:visual:ablation}
\Cref{tab:ablation} adds the slow channels and the gated residual predictor (GR) to SVReg separately and together, and \Cref{tab:ablation_gr} adds GR to the other objectives on Cube. Each part fixes a different failure of SVReg, and together they give the largest gains on Cube and PushT. On Cube, the slow channels alone give most of the gain, while the gated residual alone fails in some runs, even over a larger set of training runs, where Slow SVReg never does. In these runs the latent stays longer in an early phase of fast, jittery change and acquires cube height late. The two failing runs we continued recover with twice the training steps, so their failure is a delay beyond the step budget. On PushT, both parts raise success, but the block becomes readable mainly with the gated residual (\Cref{app:ablation}).

We find that these gains are tied to the sliced floor trained without SG. Restricting SVReg's directions to the coordinate axes gives the element-wise floor, which loses cube height and plans at about 50\% on Cube (\Cref{app:ablation}). Under SG, the predictor's default copy is a trivial solution, and SG with the gated residual collapses to a latent without cube information (\Cref{tab:ablation_gr}). Without SG, the gated residual also raises VICReg's Cube success.

\providecommand{\pend}{\textcolor{red}{pend.}}
\begin{table}[t]\small\setlength{\tabcolsep}{3pt}
\begin{minipage}[t]{0.555\linewidth}\centering
\caption{\textbf{Planning success (\%) with each part of Slow SVReg.} Hard goals on the mazes. Plain SVReg at its \Cref{tab:main} setting, the others at Slow SVReg's.}
\label{tab:ablation}
\begin{tabular}{lcccc}
\toprule
Task & plain & + GR & + Slow & + Slow + GR \\
\midrule
U-maze & 96.4\pmsd{1.3} & 98.1\pmsd{0.9} & 96.6\pmsd{1.7} & \textbf{98.7}\pmsd{0.7} \\
Medium & 81.3\pmsd{2.3} & \textbf{86.0}\pmsd{3.1} & 82.7\pmsd{2.5} & 84.7\pmsd{2.7} \\
Reacher & 98.7\pmsd{0.9} & \textbf{99.7}\pmsd{0.3} & 96.9\pmsd{1.5} & 99.6\pmsd{0.4} \\
Cube & 76.9\pmsd{8.8} & 78.6\pmsd{9.7} & 84.6\pmsd{3.3} & \textbf{89.1}\pmsd{1.6} \\
PushT & 75.0\pmsd{8.6} & 85.9\pmsd{3.9} & 78.9\pmsd{6.6} & \textbf{87.3}\pmsd{3.8} \\
\bottomrule
\end{tabular}
\end{minipage}\hfill
\begin{minipage}[t]{0.415\linewidth}\centering
\caption{\textbf{Planning success (\%) on Cube with GR added.} Each objective at its \Cref{tab:main} setting.}
\label{tab:ablation_gr}
\begin{tabular}{lcc}
\toprule
Method & plain & + GR \\
\midrule
SG + SVReg & 77.2\pmsd{1.3} & 79.9\pmsd{6.8} \\ %
SG & 80.4\pmsd{2.1} & 51.4\pmsd{7.7} \\
SG + straightening & 85.4\pmsd{1.0} & 58.6\pmsd{2.5} \\
SIGReg-BN & 87.4\pmsd{0.8} & 86.4\pmsd{0.5} \\
VICReg & 69.6\pmsd{5.8} & 78.3\pmsd{7.7} \\
\bottomrule
\end{tabular}
\end{minipage}
\end{table}

\subsection{Data Scaling}\label{sec:visual:scaling}
We test whether the gains hold beyond the dataset sizes and training budget adopted in \Cref{tab:main}.
In \Cref{fig:scaling}, Slow SVReg reaches high success from little data. On hard U-maze goals it nearly saturates with a fifth of the \Cref{tab:main} data, where SG stays far below even with all of it. On Medium and Reacher, it plans well above SVReg and SG with only 1k to 2k frames, and on Cube it plans better than both from 5k frames up to the \Cref{tab:main} size (protocol in \Cref{app:scaling}).

In \Cref{fig:stepscale}, Slow SVReg on PushT improves with longer training. With a short learning-rate cooldown, its success keeps rising up to 150k steps. SG + straightening, the next-best method, improves at a similar rate, and Slow SVReg keeps its lead.

\section{City Walks}\label{sec:city}
Inspired by the results in first-person mazes,
we next ask whether the same recipe learns a map from first-person video and actions of real places.
Wanderland~\citep{liu2026wanderland} provides 3D Gaussian-splat reconstructions~\citep{kerbl20233d} of real city blocks and buildings, captured on foot. In each scene we render first-person walks with random turns and forward steps along the capture path, and train on the frames and actions only, without poses (\Cref{app:city}). Linear readouts fit on the training walks give the agent's position and heading on held-out walks. A map score measures how well the top two principal components of the latent lay out the walked area, and a held-out disc outside the training walks tests how far the map extends.

\begin{figure}[t]
\centering
\sbox0{\includegraphics[width=\linewidth]{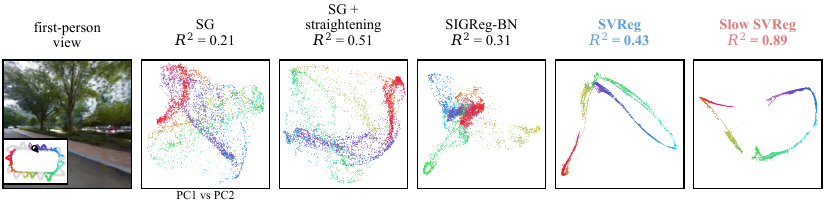}}\usebox0\plab{0.004}{0.7827}{a}\plab{0.171}{0.7827}{b}\plab{0.338}{0.7827}{c}\plab{0.506}{0.7827}{d}\plab{0.673}{0.7827}{e}\plab{0.841}{0.7827}{f}
\caption{\textbf{Held-out walks on Wanderland scene 13Skv in the top two principal components of each latent.} Color: true position. Titles: map score. (a)~A rendered first-person frame and the walked area (inset). (b--f)~SG, SG + straightening, SIGReg-BN, SVReg and Slow SVReg.}
\label{fig:city_map}
\end{figure}

\begin{table}[t]\centering\small
\caption{\textbf{Latent maps from city walks.} Scene 13Skv and the mean over ten further scenes (\Cref{tab:city_multi}). Seeds and selection in \Cref{app:city}.}
\label{tab:city}
\begin{adjustbox}{max width=\linewidth}
\begin{tabular}{lcccc@{\hspace{10pt}}cc}
\toprule
 & \multicolumn{4}{c}{scene 13Skv} & \multicolumn{2}{c}{ten further scenes (mean)} \\
\cmidrule(r{6pt}){2-5} \cmidrule{6-7}
Method & position (m) $\downarrow$ & held-out area (m) $\downarrow$ & heading ($^\circ$) $\downarrow$ & map $R^2$ $\uparrow$ & position (m) $\downarrow$ & map $R^2$ $\uparrow$ \\
\midrule
Untrained encoder (control) & 25.9 & 41.2 & 54.1 & 0.02 & 13.7 & 0.04 \\
\midrule
SG & 3.4 & 10.0 & \textbf{3.4} & 0.26 & 5.4 & 0.10 \\
SG + straightening & 9.1 & 14.5 & 11.0 & 0.24 & 5.2 & 0.21 \\
\midrule
VICReg & 12.5 & 21.3 & 21.1 & 0.12 & 5.3 & 0.13 \\
SIGReg-BN & 2.1 & 11.9 & 7.1 & 0.30 & \textbf{2.1} & 0.21 \\
\rowblue SVReg & 5.4 & 12.2 & 8.0 & 0.54 & 4.1 & 0.60 \\
\rowblue Slow SVReg (ours) & \textbf{1.6} & \textbf{6.3} & 6.3 & \textbf{0.88} & 3.1 & \textbf{0.63} \\
\bottomrule
\end{tabular}
\end{adjustbox}
\end{table}

In \Cref{tab:city}, linear readouts of the latent recover position for the objectives trained without SG. Slow SVReg reads position most accurately on scene 13Skv and remains the most accurate in the held-out area, where every readout extrapolates. SG also reads position, and it reads heading better than the SVReg variants. SG + straightening reads position in some runs and loses it in others, and among the trained models VICReg has the largest position errors. An untrained encoder of the same architecture reads position barely better than predicting the mean position and keeps no map.

The latents trained without SG also recognize places across viewing directions. For most validation frames, the nearest training frame in the latent among those facing more than $90^\circ$ away was taken within 5\,m, whereas SG mostly matches views with similar headings (\Cref{app:city}).

What Slow SVReg and SVReg add is geometry. In \Cref{fig:city_map}, the walked layout appears in the leading principal components of their latents, with no fitted readout. Slow SVReg forms the clearest map, followed by SVReg, both well above the other single-stage objectives (\Cref{tab:city}).

The map forms with experience. In \Cref{fig:walk_sweep}, position error drops by an order of magnitude as the training walks grow from two to one hundred and revisit the same streets. Earlier predictive and slow-feature models found position codes in simulated or small real environments, mostly through trained decoders (\Cref{sec:related}). Here the street layout of a reconstructed city block appears in the plain principal components of an action-conditioned JEPA trained without poses or stop-gradient.

We also train separate models on ten further scenes, four indoor and six outdoor, at the scene-13Skv setting, with two runs per scene. In \Cref{tab:city_multi}, an SVReg variant has the highest map score in all ten scenes, most often Slow SVReg. Using a linear readout, SIGReg-BN localizes best across scenes, but its leading components form no map.

\FloatBarrier
\section{Related Work}\label{sec:related}

\paragraph{Self-supervised learning and anti-collapse.}
To prevent collapse, contrastive methods use negative pairs~\citep{chen2020simple,he2020momentum}, and non-contrastive methods with stop-gradient, with or without an exponential moving average (EMA) target~\citep{grill2020bootstrap,chen2021exploring}, or centering~\citep{caron2021emerging}. These objectives are equivalent under limited assumptions~\citep{garrido2023duality}. Barlow Twins decorrelates~\citep{zbontar2021barlow}, VICReg floors element-wise standard deviations~\citep{bardes2022vicreg}, and SIGReg and VISReg match a Gaussian shape along random slices~\citep{balestriero2025lejepa,wu2026visreg}. BYOL also trains with weight-standardized group normalization instead of batch normalization~\citep{richemond2020byol}. Theory shows that the non-contrastive objective collapses without stop-gradient~\citep{tian2021understanding} and characterizes the features it learns~\citep{tang2023understanding,littwin2024jepa}. Our floor is one-sided like VICReg's and sliced like SIGReg's, and we compare a stop-gradient target with a regularized target that receives gradients by what content and geometry each keeps for a planner.

\paragraph{JEPA and world models.}
JEPA learns through temporal prediction~\citep{lecun2022path} and scales to images and video~\citep{assran2023self,bardes2024revisiting}. Latent dynamics support planning in World Models, PlaNet, Dreamer, TD-MPC and DINO-WM~\citep{ha2018world,hafner2019learning,hafner2020dream,hafner2021mastering,hansen2022temporal,hansen2024td,zhou2025dino}. PLDM trains without SG using VICReg-style losses, inverse dynamics and whole-latent slowness~\citep{sobal2025learning}, and LeWM uses SIGReg from pixels~\citep{maes2026leworldmodel}. \citet{terver2025drives} study what drives planning success in JEPA world models, and AdaJEPA adapts a pretrained JEPA world model at test time under SG~\citep{wang2026adajepa}. Content loss also appears in JEPAs that focus on slow features and fail with a fixed distractor~\citep{sobal2022joint}, and in Pong, where LeWM suppresses the ball that a stop-gradient variant recovers~\citep{karmann2026motionjepa}. We analyze how the target gradient affects content and noise, and favor slowly changing content without SG.

\paragraph{Temporal representation learning and straightening.}
Slow Feature Analysis extracts slow features under unit-variance and decorrelation constraints~\citep{wiskott2002slow}, CPC contrasts future observations with unrelated ones~\citep{oord2018representation}, and deep networks have used the temporal coherence of video frames to learn image features~\citep{mobahi2009deep,goroshin2015unsupervised,jayaraman2016slow}. Koopman methods learn linear latent dynamics, scored by VAMP~\citep{wu2020variational} or trained with autoencoders~\citep{lusch2018deep}, and JEPA connects to Koopman invariant subspaces~\citep{ruizmorales2025koopman}. Perception straightens natural-video trajectories~\citep{henaff2019perceptual}, and explicit curvature regularization supports latent planning~\citep{wang2026temporal}. Our no-stop-gradient objective is VAMP-like in the linear setting (\Cref{thm:dichotomy}), but SVReg learns smooth latent paths with nonlinear encoders and predictors, without a curvature penalty.

\paragraph{Spatial representations from prediction.}
Predictive and slow-feature models encode position, but prior work reads it mostly through trained decoders or single units. Examples are navigation agents with auxiliary tasks~\citep{mirowski2017learning,mirowski2018learning}, world models with belief states or next-frame prediction~\citep{gregor2019shaping,gornet2024automated}, and slow feature analysis of camera views in small areas~\citep{franzius2007slowness,metka2018bioinspired}. Without a decoder, coarser structure appears in the top principal components of a network that predicts 20-dimensional sensor readings~\citep{recanatesi2021predictive} and in spatially contiguous k-means clusters of a regularized world model~\citep{xia2026cloning}. In our latents, the top two principal components lay out first-person mazes and the walked streets of reconstructed city blocks, learned from frames and actions without poses.

\section{Conclusion}\label{sec:conclusion}

A latent world model for planning needs to learn \emph{content} that encodes task-relevant states, and \emph{geometry} that smoothly tracks progress toward a goal. SG, whose target is one-sided, learns content but poor geometry, and common regularizers, which constrain the latent feature distribution beyond a variance floor, also struggle with geometry. In contrast, SVReg learns smooth latent maps, but it can lose task-relevant content. To address this tradeoff, we introduce Slow SVReg, which favors slowly changing factors. It keeps this content and improves planning. Slow SVReg plans best in the hard settings of visual control tasks, and moreover, in first-person mazes and city walks, it also learns the map layout in its leading principal components.

\section*{Acknowledgments}
This research was supported by NSF Award 1922658, 2545541, AFOSR grant FA95502310139, Visko Platform, Pionex, a Google TPU Award, a Google ML and Systems Junior Faculty Award, and the Institute of Information \& Communications Technology Planning \& Evaluation (IITP) with a grant funded by the Ministry of Science and ICT (MSIT) of the Republic of Korea in connection with the Global AI Frontier Lab International Collaborative Research Agreement (No.~RS-2024-00469482 \& RS-2024-00509279). 
The compute is supported by the NYU High Performance Computing resources, services, and staff expertise. 
We gratefully acknowledge use of the research computing resources of the Empire AI Consortium, Inc, with support from Empire State Development of the State of New York, the Simons Foundation, and the Secunda Family Foundation.

\bibliographystyle{apalike}
\bibliography{ref}

\newpage
\appendix
\renewcommand{\topfraction}{0.85}

\FloatBarrier
\section{Analysis: Proofs and Discussion}\label{app:proofs}

This appendix supports the three claims of \Cref{sec:theory}. The proof of \Cref{thm:dichotomy} and its extension to actions (\Cref{app:actions}) show that without stop-gradient the loss penalizes unpredictable variance (\Cref{sec:dichotomy}). \Cref{app:prop-discussion} shows that with finite data a stop-gradient encoder can read noise while its latent statistics look healthy (\Cref{sec:collapse}). \Cref{app:content} shows how a hard-to-predict factor loses a latent coordinate under limited capacity (\Cref{sec:content}). \Cref{app:thm-discussion} relates the analysis to prior work.

\subsection{Setting and Notation}

For \Cref{thm:dichotomy} and its action extension (\Cref{app:actions}), we use a linear encoder $\Phi(o) = Wo$ with $W \in \R^{r \times d}$, a linear predictor $A \in \R^{r \times r}$, mean-centered observations and the sample covariances $\hat\Sigma_{00} = \tfrac{1}{T-1}\sum_t \bm{o}_t \bm{o}_t^\top$, $\hat\Sigma_{11} = \tfrac{1}{T-1}\sum_t \bm{o}_{t+1}\bm{o}_{t+1}^\top$, $\hat\Sigma_{01} = \tfrac{1}{T-1}\sum_t \bm{o}_t \bm{o}_{t+1}^\top$. Index $0$ refers to the current frame and index $1$ to the next frame. Write $\Sigma_z = W\hat\Sigma_{00}W^\top$ and $C_z = W\hat\Sigma_{01}W^\top$, and assume $\Sigma_z$ is invertible, so that no latent direction is constant on the training frames. We treat each block of $T-1$ frames as centered, so these second-moment matrices are covariances.

\subsection{Proof of \Cref{thm:dichotomy}}

\paragraph{Part 1 (without stop-gradient).} The empirical loss expands as
\[
  \mathcal{L}(W,A) \;=\; \tfrac{1}{T-1}\sum_t \|A W \bm{o}_t - W \bm{o}_{t+1}\|^2
  \;=\; \mathrm{tr}(A\Sigma_z A^\top) - 2\,\mathrm{tr}(A C_z) + \mathrm{tr}\!\bigl(W\hat\Sigma_{11}W^\top\bigr),
\]
which is quadratic in $A$. Setting $\nabla_A \mathcal{L} = 2A\Sigma_z - 2C_z^\top = 0$ gives $A^\star = C_z^\top\Sigma_z^{-1}$. Since $\Sigma_z$ is positive definite, the loss is strictly convex in $A$ and $A^\star$ is its unique minimizer. With $A^{\star\top} = \Sigma_z^{-1}C_z$, the first two terms become $\mathrm{tr}(C_z^\top\Sigma_z^{-1}C_z) - 2\,\mathrm{tr}(C_z^\top\Sigma_z^{-1}C_z)$, so
\[
  \mathcal{L}^\star_{\mathrm{no\text{-}sg}}(W) \;=\; \mathrm{tr}\!\bigl(W\hat\Sigma_{11}W^\top\bigr) \;-\; \mathrm{tr}(C_z^\top \Sigma_z^{-1} C_z).
\]
This is the trace of the residual covariance of $W\bm{o}_{t+1}$ after linear regression on $\bm{z}_t$, the unexplained target variance.

\paragraph{Part 2 (with stop-gradient, frozen $W_{\mathrm{tgt}}$).} The target is $z'_{t+1} = W_{\mathrm{tgt}}\,\bm{o}_{t+1}$, treated as constant under $\nabla_W$. Let $\hat\Sigma_{zz'} = W\hat\Sigma_{01}W_{\mathrm{tgt}}^\top$ and $\Sigma_{z'z'} = W_{\mathrm{tgt}}\hat\Sigma_{11}W_{\mathrm{tgt}}^\top$. The loss is
\[
  \mathcal{L}(W,A) \;=\; \mathrm{tr}(A\Sigma_z A^\top) \;-\; 2\,\mathrm{tr}(A\,\hat\Sigma_{zz'}) \;+\; \mathrm{tr}(\Sigma_{z'z'}).
\]
Setting $\nabla_A\mathcal{L} = 0$ gives $A^\star = \hat\Sigma_{zz'}^\top\Sigma_z^{-1}$. Substituting as in Part~1 gives
\[
  \mathcal{L}^\star_{\mathrm{sg}}(W;\,W_{\mathrm{tgt}}) \;=\; \mathrm{tr}(\Sigma_{z'z'}) \;-\; \mathrm{tr}\!\bigl(\hat\Sigma_{zz'}^\top\Sigma_z^{-1}\hat\Sigma_{zz'}\bigr).
\]
The first term does not depend on $W$, so minimizing the loss over $W$ maximizes the second term, the variance of the frozen target predicted from $\bm{z}_t$. In both parts $\nabla_A\mathcal{L}=0$ at $A^\star$, so by the envelope theorem the gradient of $\mathcal{L}^\star$ in $W$ equals the gradient of $\mathcal{L}$ at $A^\star$, the encoder update for a predictor kept at its least-squares fit. \qed

\subsection{Action-Conditioned Predictor}\label{app:actions}

With the predictor $\hat{\bm{z}}_{t+1} = A\bm{z}_t + B\bm{a}_t$, both parts of \Cref{thm:dichotomy} keep their form once the action is linearly regressed out of the observations. The reason is that the optimal $B$ predicts the part of the next latent that the action explains linearly. Without stop-gradient, the loss then penalizes only the next-step variance left unexplained by the current latent and action together. The action predicts most of the agent's own motion, so this motion adds little to the penalty, while contact-driven object motion adds to it. This is why \Cref{sec:content} defines the residual $\delta$ given both the state and the action.

\paragraph{Residual statistics.} Let the actions $\bm{a}_t \in \R^m$ be mean-centered together with the observations, with invertible sample covariance $\hat\Sigma_{aa} = \tfrac{1}{T-1}\sum_t \bm{a}_t\bm{a}_t^\top$ and cross-covariances $\hat\Sigma_{0a} = \tfrac{1}{T-1}\sum_t \bm{o}_t\bm{a}_t^\top$ and $\hat\Sigma_{1a} = \tfrac{1}{T-1}\sum_t \bm{o}_{t+1}\bm{a}_t^\top$ over the same $T-1$ pairs. For each pair $t$, the least-squares residuals of the current and next frame after regressing out the action $\bm{a}_t$ are
\[
  \tilde{\bm{o}}_{0,t} = \bm{o}_t - \hat\Sigma_{0a}\hat\Sigma_{aa}^{-1}\bm{a}_t,
  \qquad
  \tilde{\bm{o}}_{1,t} = \bm{o}_{t+1} - \hat\Sigma_{1a}\hat\Sigma_{aa}^{-1}\bm{a}_t,
\]
and their sample covariances are, for $i, j \in \{0, 1\}$,
\[
  \hat\Sigma_{ij|a} = \hat\Sigma_{ij} - \hat\Sigma_{ia}\hat\Sigma_{aa}^{-1}\hat\Sigma_{aj},
\]
with $\hat\Sigma_{ai} = \hat\Sigma_{ia}^\top$ and $\hat\Sigma_{10} = \hat\Sigma_{01}^\top$.

\paragraph{Without stop-gradient.} For fixed $(W, A)$, the loss $\mathcal{L}(W, A, B) = \tfrac{1}{T-1}\sum_t \|AW\bm{o}_t + B\bm{a}_t - W\bm{o}_{t+1}\|^2$ is a convex quadratic in $B$, minimized by $B^\star(A) = (W\hat\Sigma_{1a} - AW\hat\Sigma_{0a})\hat\Sigma_{aa}^{-1}$. With this choice the prediction error is
\[
  AW\bm{o}_t + B^\star(A)\,\bm{a}_t - W\bm{o}_{t+1} = AW\tilde{\bm{o}}_{0,t} - W\tilde{\bm{o}}_{1,t},
\]
so $\min_B \mathcal{L}(W, A, B)$ is the action-free loss of Part~1 evaluated on the residuals. Minimizing it over $A$ as in Part~1, with $\tilde\Sigma_z = W\hat\Sigma_{00|a}W^\top$ and $\tilde C_z = W\hat\Sigma_{01|a}W^\top$, gives $A^\star = \tilde C_z^\top\tilde\Sigma_z^{-1}$, $B^\star = B^\star(A^\star)$ and
\[
  \mathcal{L}^\star_{\mathrm{no\text{-}sg}}(W) = \mathrm{tr}\!\bigl(W\hat\Sigma_{11|a}W^\top\bigr) - \mathrm{tr}\!\bigl(\tilde C_z^\top\tilde\Sigma_z^{-1}\tilde C_z\bigr).
\]
Minimizing over $B$ and then over $A$ gives the joint minimum over $(A, B)$. We assume $\tilde\Sigma_z$ is invertible, which holds exactly when the predictor input $(\bm{z}_t, \bm{a}_t)$ has an invertible sample covariance. This excludes latent directions that are exact linear functions of the action on the sample.

\paragraph{With stop-gradient.} With the frozen target $W_{\mathrm{tgt}}\bm{o}_{t+1}$, this elimination gives
\[
B^\star(A) = (W_{\mathrm{tgt}}\hat\Sigma_{1a} - AW\hat\Sigma_{0a})\hat\Sigma_{aa}^{-1}
\]
and the error $AW\tilde{\bm{o}}_{0,t} - W_{\mathrm{tgt}}\tilde{\bm{o}}_{1,t}$. Part~2 therefore holds with $\Sigma_z$ replaced by $\tilde\Sigma_z$, $\hat\Sigma_{zz'} = W\hat\Sigma_{01|a}W_{\mathrm{tgt}}^\top$ and the constant first term $\mathrm{tr}(W_{\mathrm{tgt}}\hat\Sigma_{11|a}W_{\mathrm{tgt}}^\top)$.

\subsection{Signal Collapse (Proposition~\ref{prop:stable})}\label{app:prop-discussion}

\paragraph{Observation model.} Observations combine a zero-mean $k$-dimensional signal with independent noise, $\bm{o}_t = U\bm{s}_t + \sigma\bm{n}_t$, where $U\in\R^{d\times k}$ has orthonormal columns, $\sigma>0$, and $\bm{n}_t\sim\mathcal{N}(0,I_d)$ is independent across time and of the signal (\Cref{app:toys}).

\paragraph{Spurious sample structure in finite data.} For $T$ samples of i.i.d.\ noise $n_t \in \R^d$, random-matrix theory gives the empirical lag-1 cross-covariance $\tfrac{1}{T-1}\sum_t n_t n_{t+1}^\top$ a nonzero spectral norm despite its zero population value. Its scale is of order $\sqrt{d/T}$ for $d \ll T$ and grows further for $d \gtrsim T$, contributing spurious structure of order $\sigma^2\sqrt{d/T}$ to the observation lag-covariance.

The fixed-target objective in Part~(2) can select this spurious structure when it exceeds the signal's lag structure. An encoder can keep healthy latent variance while reading noise instead of state. We call this \emph{signal collapse}, and latent statistics can hide it.

\begin{proposition}[Latent covariance does not identify signal recovery]\label{prop:stable}
Let the signal have zero mean and unit covariance, and the noise be isotropic with positive variance. Let the faithful encoder $\Phi^\star$ project onto the signal directions $U$ and the collapsed encoder $\tilde\Phi$ onto $k$ directions orthogonal to $U$, both followed by the same output map with orthonormal columns. Then the two population latent covariances differ only by a global scale, which a variance regularizer can hide. A linear probe of the true state separates them, with population $R^2$ equal to $1/(1+\sigma^2)$ at $\Phi^\star$ and $0$ at $\tilde\Phi$.
\end{proposition}

\paragraph{Proof.}

Let $\Phi^\star(o) = M_s U^\top o$ with $M_s \in \R^{r\times k}$ satisfying $M_s^\top M_s = I_k$, and $\tilde\Phi(o) = M_s V^\top o$ with $V \in \R^{d\times k}$ orthonormal and $V^\top U = 0$ (assuming $d \ge 2k$).

\paragraph{Part 1: Covariance shape.} With $\Sigma_o = U\,\mathrm{Cov}(s)\,U^\top + \sigma^2 I_d$ and $\mathrm{Cov}(s) = I_k$:
\[
\begin{aligned}
  \mathrm{Cov}(\Phi^\star(o)) &= M_s U^\top \Sigma_o U M_s^\top = (1+\sigma^2)\,M_s M_s^\top, \\
  \mathrm{Cov}(\tilde\Phi(o)) &= M_s V^\top \Sigma_o V M_s^\top = \sigma^2\,M_s M_s^\top.
\end{aligned}
\]
Both are positive multiples of the same matrix, so the two encoders share their eigenvalue profile up to a global scale, their scale-invariant variance ratios, and their Pearson correlations. Rescaling the collapsed encoder by $\sqrt{1+\sigma^2}/\sigma$ makes its population covariance identical to the faithful encoder's, so covariance alone cannot identify signal recovery when the encoder gain is unrestricted.

\paragraph{Part 2: Ground-truth probe.} Define the population coefficient of determination $R^2(z, s) = 1 - \E\|s - \hat s(z)\|^2 / \E\|s\|^2$, where $\hat s(z)$ is the optimal linear predictor of $s$ from $z$. For $\tilde\Phi$, $z = \sigma M_s V^\top n$ contains no $s$ component, so $R^2(\tilde\Phi(o), s) = 0$. For $\Phi^\star$, $z = M_s s + \sigma M_s U^\top n$ and, since $M_s^\top M_s = I_k$, applying $M_s^\top$ gives the equivalent observation model $M_s^\top z = s + \sigma U^\top n$, each signal component observed with signal variance $1$ and noise variance $\sigma^2$, so $R^2(\Phi^\star(o), s) = 1/(1+\sigma^2)$ per component. The gap $1/(1+\sigma^2)$ is positive at any finite noise level. On the training trajectory, finite-sample $R^2$ can exceed this ceiling when $d \ge T$ through memorization, which is why the population gap is measured on held-out trajectories. \qed

\subsection{Content Competition (Proposition~\ref{prop:content})}\label{app:content}

Proposition~\ref{prop:content} explains why training without stop-gradient can drop a hard-to-predict factor, such as the cube's height in \Cref{sec:content}. It uses the scalar mode of \Cref{sec:content}, $s_{t+1}=\varphi s_t+\kappa a_t+\epsilon_t$ with $|\varphi|<1$, innovation (process noise) variance $q=\mathrm{Var}(\epsilon_t)$, $\kappa^2+q>0$, and stationary variance $v=\mathrm{Var}(s_t)$. Actions and innovations have zero mean, so the stationary mode has zero mean. Its \emph{one-step residual} $\delta=q/v$ is the share of the mode's variance that is not predicted by the latent and the action. For a unit-variance action, the mode's variance splits into shares $\varphi^2$ from persistence, $\alpha=\kappa^2/v$ from the action and $\delta$ from the innovation, which sum to one, so $1-\delta$ is the $R^2$ of the best linear one-step prediction.

\begin{proposition}[Competition for one latent coordinate]\label{prop:content}
Let two independent stationary modes $i$ and $j$, each driven by its own action, have persistence $\varphi_i,\varphi_j$, stationary variances $v_i,v_j$ and one-step residuals $\delta_i,\delta_j$. They share a one-dimensional latent, trained by gradient flow with encoder learning rate $\eta$, a predictor held at its least-squares fit, and SVReg. In one dimension, SVReg is the floor $\lambda(1-c)_+^2$ with $\lambda>0$, where $c$ is the standard deviation of the latent. Start from the endpoint where the latent holds only mode $i$ at its equilibrium scale, $c=\lambda/(\lambda+\delta_i)$ without stop-gradient and any $c\ge1$ with it. To first order, a small encoder component along mode $j$ grows at rate $\gamma$, with
\[
  \gamma_{\mathrm{no\text{-}sg}} = 2\eta v_j\bigl[\delta_i - \delta_j - (\varphi_i-\varphi_j)^2\bigr],
  \qquad
  \gamma_{\mathrm{sg}} = 2\eta v_j\,\varphi_i(\varphi_j-\varphi_i).
\]
\end{proposition}

\paragraph{Comparing the rates.} Substituting $\delta = 1-\varphi^2-\alpha$ gives
\[
  \gamma_{\mathrm{no\text{-}sg}} = 2\gamma_{\mathrm{sg}} + 2\eta v_j(\alpha_j-\alpha_i).
\]
Both rates contain the persistence term $\varphi_i(\varphi_j-\varphi_i)$, which for $\varphi_i>0$ favors a more persistent newcomer. Only training without stop-gradient adds the action term, so only there can a less persistent, action-driven mode displace a passive, persistent one. Without stop-gradient, the brackets for $j$ entering $i$ and for $i$ entering $j$ sum to $-2(\varphi_i-\varphi_j)^2$, so both endpoints can be stable and the first mode to fill the latent can keep it. With stop-gradient and distinct positive persistences, only the more persistent endpoint is stable.

\paragraph{Proof.} Actions (unit variance) and innovations have zero mean, are fresh at each step, and are independent of each other and of the state, so $v = (\kappa^2+q)/(1-\varphi^2)$. For one mode with encoder $z_t = ws_t$ ($w \neq 0$), scale $c = |w|\sqrt{v}$ and predictor $\hat z_{t+1} = \hat\varphi z_t + \hat\kappa a_t$, the one-step prediction error $e$ splits into uncorrelated state, action and innovation terms, so
\[
  \E e^2 = c^2\bigl[(\hat\varphi-\varphi)^2 + \delta\bigr] + (\hat\kappa-w\kappa)^2 .
\]
Write $h(c) = \lambda(1/c-1)_+$ for the floor's push, so that $\partial_w \lambda(1-c)_+^2 = -2h(c)\,wv$.

\emph{Endpoint.} Without stop-gradient, the loss at the fitted predictor ($\hat\varphi=\varphi$, $\hat\kappa=w\kappa$) is $\delta c^2 + \lambda(1-c)_+^2$, where $\delta c^2$ is $\mathcal{L}^\star_{\mathrm{no\text{-}sg}}$ for this mode (\Cref{app:actions}). The predictor's gradient vanishes at its fit, so by the envelope theorem the encoder gradient is the derivative of this reduced loss. For $\delta>0$ it is strictly convex with its minimum where $h(c)=\delta$, at $c = \lambda/(\lambda+\delta)$. So the latent settles slightly below the floor, and the floor keeps pushing for more variance with strength $\delta$. With stop-gradient, $w$ in the target $ws_{t+1}$ is held fixed, the encoder gradient of $\E e^2$ is $2\hat\varphi(\hat\varphi-\varphi)\,wv$, which vanishes at the fitted predictor, and any $c \ge 1$, where $h=0$, is a fixed point. So the latent can sit at or above the floor, where the floor stops pushing.

\emph{Growth rate.} Perturb the encoder to $z = w s_i + \zeta s_j$, with $c_i = |w|\sqrt{v_i}$. For independent modes, the fitted $\hat\varphi$ and the latent standard deviation change only at $O(\zeta^2)$, and the loss splits by mode. The mode-$j$ part is $\zeta^2 v_j[(\hat\varphi -\varphi_j)^2 + \delta_j]$, and the floor contributes $-2h\,\zeta v_j$ to the gradient in $\zeta$. With stop-gradient, only the prediction is differentiated, and the prediction-loss gradient is $2\hat\varphi\,\E[e\,s_j] = 2\hat\varphi(\hat\varphi-\varphi_j)\,\zeta v_j$. At $\hat\varphi=\varphi_i$,
\[
  \text{no SG: } \dot\zeta = 2\eta v_j\,\zeta\bigl[h(c_i) - \delta_j - (\varphi_i-\varphi_j)^2\bigr], \qquad
  \text{SG: } \dot\zeta = 2\eta v_j\,\zeta\bigl[h(c_i) + \varphi_i(\varphi_j - \varphi_i)\bigr],
\]
up to $O(\zeta^2)$. With $h(c_i)=\delta_i$ without stop-gradient and $h(c_i)=0$ with it, these are the rates of the proposition. In words, without stop-gradient a new mode enters when the floor's push $\delta_i$ outweighs its cost, its own unpredictable variance $\delta_j$ plus $(\varphi_i-\varphi_j)^2$ for sharing one predictor coefficient. With stop-gradient, it enters when it is more persistent than the incumbent, for $\varphi_i>0$. \qed

\paragraph{Shrinkage alone keeps the mode.} For any $\lambda>0$, the endpoint scale without stop-gradient is $\lambda/(\lambda+\delta) \ge \lambda/(\lambda+1)$, since $\delta \le 1$, which is $0.5$ at $\lambda=1$ and $0.99$ at $\lambda=100$. The linear readout $z/w$ recovers $s$ at any nonzero scale, so losing a mode needs another mechanism, such as competition for a limited latent.

\subsection{Relation to Prior Work}\label{app:thm-discussion}

\paragraph{CCA and VAMP.} Part~(1) of \Cref{thm:dichotomy} restates known VAMP and Koopman results~\citep{mardt2018vampnets,wu2020variational,ruizmorales2025koopman}. With whitened target latents ($W\hat\Sigma_{11}W^\top = I_r$), its loss is $r-\sum_i\rho_i^2$, where $\rho_i$ are the canonical correlations between $W\bm{o}_t$ and $W\bm{o}_{t+1}$. Minimizing the loss then maximizes the VAMP-2 score~\citep{wu2020variational}, with one $W$ shared across both time steps.

\paragraph{Per-mode analyses.} Earlier analyses follow each mode separately. They derive contraction without stop-gradient for static views~\citep{tian2021understanding} and subspace dynamics under stop-gradient with an optimal predictor~\citep{tang2023understanding,ni2024bridging}. Under stop-gradient, joint updates of encoder and predictor generally do not follow the gradient of a scalar objective~\citep{ponce2026dual}, so Part~(2) of \Cref{thm:dichotomy} describes a single update against a frozen target. Proposition~\ref{prop:content} adds the action-conditioned residual $\delta$ and the equilibrium scale $\lambda/(\lambda+\delta)$, where the floor balances the prediction loss.

\section{Linear Dynamics Experiment Details}\label{app:toys}
\begin{figure}[t]
\centering
\sbox0{\includegraphics[width=\linewidth]{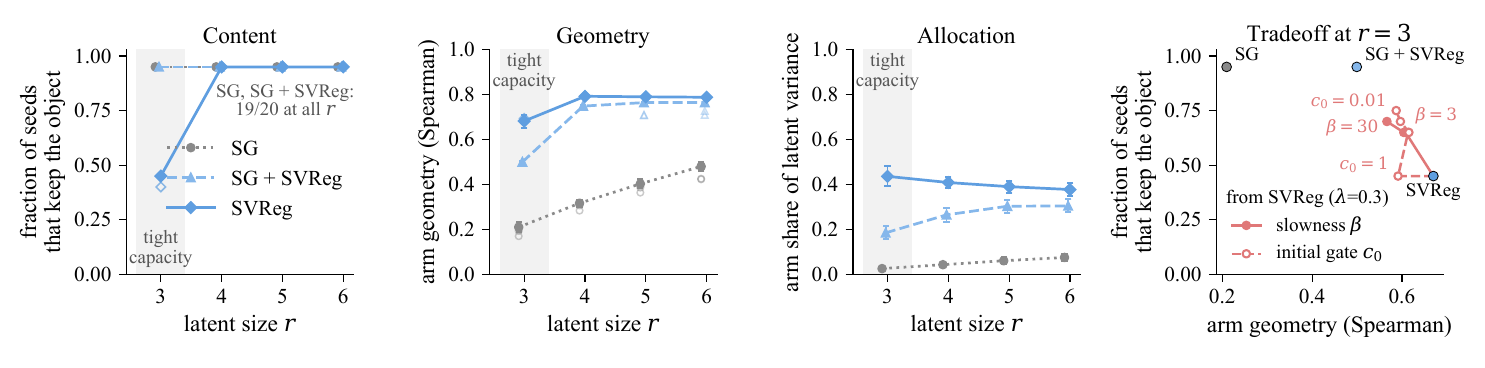}}\usebox0\plab{0.0056}{0.975}{a}\plab{0.2732}{0.975}{b}\plab{0.5182}{0.975}{c}\plab{0.749}{0.975}{d}
\caption{\textbf{Object content and arm geometry under limited latent capacity.} (a--c)~Objects kept, arm geometry and arm share of latent variance against latent size. (d)~Latent size 3 with varied slowness weight and initial gate.}
\label{fig:toy_content}
\end{figure}

\label{sec:setup}
\paragraph{High-dimensional linear dynamics.} A low-dimensional state $\bm{s}_t \in \R^k$ follows $\bm{s}_{t+1} = R\,\bm{s}_t + K\bm{a}_t + \bm{\epsilon}_t$, where $R \in \R^{k \times k}$ is the transition matrix, $K \in \R^{k \times m}$ maps actions to states, and $\bm{\epsilon}_t$ is a zero-mean innovation. The geometry experiment of \Cref{sec:toys} uses deterministic dynamics without actions. The agent observes the state through a random orthonormal map $U \in \R^{d \times k}$ with isotropic Gaussian noise:
\[
  \bm{o}_t = U \bm{s}_t + \sigma n_t, \qquad n_t \sim \mathcal{N}(0, I_d), \quad t = 0, 1, \ldots, T-1.
\]
The noise level $\sigma$ and sample size $T$ govern finite-sample uncertainty.

The linear world tests the geometry and content predictions of \Cref{sec:theory}. Both experiments sample training windows i.i.d.\ from a fixed trajectory, as in the visual experiments.

\paragraph{Geometry: recovering the dynamics.}
A state following the model above traces a circle at frequency $\omega$, observed through a random linear map with noise $\sigma{=}0.3$. Recovering this circle on new episodes allows latent distance to follow the state. We sweep $\omega$ from slow to fast and the trajectory length $T$ from 250 to 2{,}000 frames. We compare SG, SVReg and SG + SVReg over eight training runs, with each weight $\lambda$ selected on the other runs. The held-out probe measures signal recovery on a fresh episode, with an oracle ceiling set by observation noise.

\paragraph{Content: an object under limited latent size.}
An action-driven arm and a contact-driven object, each with a 2-D position, compete for latent capacity. The object jumps to a random position when the arm enters a contact radius. Observations are a fixed random linear map of the state plus noise, as above, and a linear encoder with an action-conditioned linear predictor learns a latent of size $r$. At $r{=}3$, the latent is smaller than the 4-D state. Over 20 runs, we count how often the object is kept (held-out $R^2 \geq 0.7$) and measure \emph{arm geometry}, the rank correlation of latent and arm distances on held-out frames. Open markers show variants with 3 and 6 distractors.

At $r{=}3$, the latent cannot represent both factors fully, so each method trades one against the other. In \Cref{fig:toy_content}, SVReg keeps the arm and drops the object in about half the runs, while SG and SG + SVReg keep the object at the cost of arm geometry. With more capacity, all three lose the object in at most one run, and SG still spends little latent variance on the arm. Across latent sizes, SVReg has the best arm geometry and SG the worst. A larger slowness weight or a smaller initial gate moves SVReg along this frontier toward keeping the object (\Cref{fig:toy_content}d). The slow channel here is one of three latent dimensions, a much larger share than in the visual models.

\section{Experimental Setup Details}\label{app:setup}

\subsection{Data and Baselines}
\paragraph{Data.} PointMaze U-maze (5k frames), PointMaze Medium (20k frames) and Reacher (5k frames) are each trained on one exploration trajectory with uniform-random actions. Cube is trained on 2{,}000 episodes of the OGBench scripted expert (80k frames). PushT uses the offline dataset of DINO-WM~\citep{zhou2025dino}. In first-person PointMaze, the agent drives a unicycle through the maze and sees only the corridor in front of it. Its data come from one exploration trajectory of smooth random driving, with 5k frames on the U-maze and 20k on Medium, and the encoder and planner follow the top-down setup.

\paragraph{Baselines.} VICReg applies its variance and covariance terms to all frames of the batch. The straightening penalty skips near-stationary steps and averages over the remaining ones. Two-stage selects its SG trunk and head learning rates by the selection rule of the other methods.

\subsection{World Model Architecture and Training}\label{app:arch}
\paragraph{Encoders.} \Cref{tab:arch} lists the encoder of each task. The methods on a task share it, so within a column of \Cref{tab:main} they differ only in their objective, and Slow SVReg also in its gated residual predictor. PushT initializes GeM pooling at $p{=}1$ (average pooling).

\begin{table}[!htbp]\centering\small
\caption{\textbf{World models and training per task.} One frame is five environment steps.}
\label{tab:arch}
\newcolumntype{L}[1]{>{\raggedright\arraybackslash}p{#1}}
\begin{adjustbox}{max width=\linewidth}
\begin{tabular}{@{}l L{2.0cm} L{2.0cm} L{2.0cm} L{2.4cm} L{2.4cm}@{}}
\toprule
 & U-maze & Medium & Reacher & Cube & PushT \\
\midrule
\multicolumn{6}{@{}l}{\emph{Data and model}} \\
Data & \multicolumn{3}{l}{one uniform-random exploration trajectory} & OGBench scripted expert & DINO-WM offline data \\
Training frames & 5k & 20k & 5k & 80k & 470k \\
Image size & 224 & 224 & 128 & 224 & 224 \\
Encoder & \multicolumn{3}{l}{ResNet (5 blocks), BatchNorm} & \multicolumn{2}{l}{pre-activation ResNet (5 blocks), GroupNorm} \\
Readout & flatten & flatten & global average & GeM & GeM \\
Latent dim.\ $D$ & 128 & 128 & 128 & 384 & 128 \\
Predictor & \multicolumn{5}{l}{block-causal ViT (depth 6), 3-frame history, 10-d action embedding concatenated to the latent} \\
\midrule
\multicolumn{6}{@{}l}{\emph{Training}} \\
Steps $\times$ batch & 10k $\times$ 32 & 20k $\times$ 32 & 50k $\times$ 32 & 50k $\times$ 64 & 50k $\times$ 32 \\
Optimizer & \multicolumn{5}{l}{Adam (encoder), AdamW with lr $5{\times}10^{-4}$ (predictor), no schedule} \\
Slow SVReg & \multicolumn{5}{l}{slow channels: last $1/8$ of the latent, weight $1/\lambda$; gated residual: gate initialized at $c_0 = 0.1$} \\
Slow SVReg ($\lambda$, lr) & 100, $10^{-4}$ & 10, $10^{-3}$ & 1, $10^{-5}$ & 30, $10^{-4}$ & 100, $10^{-4}$ \\
Two-stage & \multicolumn{5}{l}{frozen SG encoder, Linear--GELU--Linear head trained with SVReg} \\
\midrule
\multicolumn{6}{@{}l}{\emph{Planning (CEM)}} \\
Samples / elites / iters & 200 / 30 / 30 & 200 / 30 / 30 & 200 / 30 / 30 & 300 / 30 / 10 & 200 / 30 / 30 \\
Horizon (frames) & 10 (easy 5) & 10 (easy 5) & 10 & 5 & 5 \\
Replan every / step budget & 1 frame / 100 & 1 frame / 100 & 1 frame / 100 & 5 frames / 50 & 1 frame / 100 \\
\bottomrule
\end{tabular}
\end{adjustbox}
\end{table}

\paragraph{Normalization.} In pilots, SG with group normalization failed on the mazes, and batch normalization lowered SVReg on PushT. The mazes and Reacher therefore use batch normalization, and Cube and PushT use group normalization.

\subsection{Planning}\label{app:bounds}
\paragraph{Planner.} The planner is CEM model-predictive control. On the mazes and Reacher, its cost is the squared L2 latent distance to the goal summed over all predicted steps, with weight $2^t$ on step $t$. PushT uses the final predicted step for the early CEM iterations and this weighted cost afterwards, and Cube uses the final predicted state only. Sampled actions are clamped to $[-1,1]$ on the mazes and Reacher. On PushT, unbounded CEM drives the agent out of the arena, so actions are clamped to the data's action range, $\pm 4$ in normalized units or about the 0.1--99.9th percentiles. The Cube planner is unbounded, and OGBench clips Cube actions to $[-1,1]$. Bounding it lowers SG and SVReg by a few points and keeps their order.

\paragraph{Goals.} In PointMaze, easy goals are 25 steps from the start with no filter on how far they lie, and hard goals are 50 steps away and at least 3 maze units apart. Cube goals are 25 environment steps ahead with a 50-step budget, as in LeWM~\citep{maes2026leworldmodel}. The Cube \emph{all} column scores all goals, and the $\geq$4\,cm column scores only goals whose cube moves more than 4\,cm (about 60\% of goals). Our start and goal frames come from held-out episodes, whereas LeWM draws them from its training episodes.

\paragraph{Scoring.} \emph{Planning success} is the percentage of episodes that reach the goal within the step budget (0 to 100, higher is better). Each trained model is scored by its mean success over three planning seeds, with 50 episodes each or 100 on Reacher. The plan column of the geometry tables in \Cref{app:pca} reports success at the \Cref{tab:main} setting, on hard goals in PointMaze.

\subsection{Hyperparameter Selection}
\paragraph{Grids.} Encoder learning rates range over $\{10^{-6}, 10^{-5}, 10^{-4}, 5{\times}10^{-4}, 10^{-3}\}$. Regularization weights range from 0.3 to 1000 for SVReg and Slow SVReg, 1 to 300 for SG + SVReg, 0.01 to 100 for VICReg and $10^{-4}$ to 1000 for SG + straightening. SIGReg-BN uses 0.3 to 30 with pooled statistics and 0.05 to 192 with per-frame statistics. For two-stage, the SG trunk is trained with learning rate $10^{-6}$ or $10^{-5}$, and the SVReg head with learning rate $10^{-4}$, $5{\times}10^{-4}$ or $10^{-3}$. The head's SVReg weight is 1 on Cube and 10 on the other tasks.

\paragraph{Selection rule.} Settings are selected on a separate planning seed. Grids are trained on three training seeds, and the leading settings on three more. For each method and task we take the setting with the highest mean success over all six, breaking ties by the lower learning rate and then the smaller weight. Each column is selected on its own goals, so the easy, hard, all and $\geq$4\,cm columns can use different settings.

\paragraph{SIGReg-BN.} SIGReg-BN screens its grid on one training seed and trains its two best settings per task on the rest. The per-frame grid multiplies the pooled setting's weight by 1/64, 1/16, 1/4, 1 and 4 at three learning rates around the pooled setting's, and extends an axis while the best setting lies on its edge. With pooled statistics, PushT uses encoder learning rate $10^{-5}$ only. The statistic form omitted from \Cref{tab:main} plans worse on all tasks.

\paragraph{First-person PointMaze.} Each method is selected from its learning-rate grid as in \Cref{tab:main}. SIGReg-BN (both statistic forms), VICReg and SG + SVReg also screen the weight grids of their top-down maze runs and extend their best settings before selection.

\subsection{Metrics}
\paragraph{Readouts (content).} Readout $R^2$, AUC, distance rank correlation, jaggedness and the participation ratio (eff.\ dim) are defined in \Cref{sec:measures}. Readouts are ridge regressions ($\alpha{=}1$) fit on one set of held-out episodes and scored on another, with 18 and 7 episodes on Cube and 15 and 6 on PushT.
\begin{itemize}[leftmargin=*,itemsep=1pt,topsep=2pt]
\item \textbf{Cube.} \emph{Contact AUC} detects gripper--cube contact with a linear readout. \emph{Free cube $R^2$} reads the cube position on frames without contact and with the gripper at least 5\,cm away, and \emph{grasped arm $R^2$} reads the gripper position on contact frames. The cube-height $R^2$ in \Cref{fig:cube_height}b also comes from a linear readout.
\item \textbf{PushT.} \emph{Agent $R^2$} reads the agent position. \emph{Free block $R^2$} reads the block position and angle ($x$, $y$, $\cos\theta$, $\sin\theta$) on frames where the block does not move and the agent is at least 60\,px away, and \emph{pushed block $R^2$} reads them on frames where the block moves.
\end{itemize}

\paragraph{Distance correlations (geometry).} Higher is better for all measures below, and rank correlations range from $-1$ to $1$.
\begin{itemize}[leftmargin=*,itemsep=1pt,topsep=2pt]
\item \textbf{Reacher.} \emph{Dist.\ $\rho$} is the Spearman correlation of latent and joint-space distance. \emph{10-NN} is the overlap of the 10 nearest neighbors of a frame in latent and joint space (0 to 1).
\item \textbf{PointMaze.} \emph{Geodesic $\rho$} is the Spearman correlation between the latent L2 distance of two held-out frames and the shortest-path distance between their agent positions inside the maze. It uses 150 anchors $\times$ 1{,}500 frames from the second half of the held-out trajectory. \emph{Local $\rho$} restricts it to pairs at most one maze cell apart.
\item \textbf{Cube and PushT.} \emph{Cost-to-goal $\rho$} is the Spearman correlation, over 20{,}000 random pairs of held-out frames, between the latent L2 distance (the planner's cost) and the true state distance. The true state is the gripper and cube positions in meters on Cube (all 25 held-out episodes). On PushT, it is the agent and block positions in pixels plus the block-angle difference scaled by $20/(\pi/9)$\,px/rad, over all 21 held-out episodes. This scaling gives the angle tolerance of the success check, $\pi/9$, the weight of its position tolerance, 20\,px.
\end{itemize}

\paragraph{Jaggedness.} Jaggedness uses a 9-step moving average over the full latent. In PointMaze it is the mean over five smooth shortest routes, a fixed route and four routes between random cell pairs at least half as long. The routes are rendered in the simulator at zero velocity, with first-person views facing the direction of travel. The Reacher sweep turns the shoulder through $360^\circ$ at elbow $60^\circ$. Cube has no scripted motion, so its jaggedness is the mean over each held-out episode's frames before the first gripper--cube contact, over 25 segments with at least 20 frames and 2\,cm of arm travel each. The true state path on these segments has jaggedness 1.09.

\subsection{Figures}

\paragraph{Cube content and geometry (\Cref{fig:cube_height}).} Each curve in panel (a) varies one knob, either SVReg predictor depth (1 to 6 blocks), SVReg slowness weight $\beta$, or the gate of the gated residual predictor, held fixed at $c_0$, without and with slow channels ($\beta=1$, $\lambda=10$). \emph{Body $\rho$} is the Spearman correlation between the latent L2 distance and the true effector distance over held-out frame pairs. The black-edged circle marks Slow SVReg at its \Cref{tab:main} setting, and panel (b) uses the \Cref{tab:main} settings. The gate curves use runs in which the gated residual alone trains well.

\section{Latent Geometry and Content}\label{app:pca}
PCA panels show the top two principal components of held-out frames from the \Cref{tab:main} checkpoints, and the tables use the full latent.
\paragraph{PointMaze.}
\begin{figure}[p]
\centering
\includegraphics[width=\textwidth]{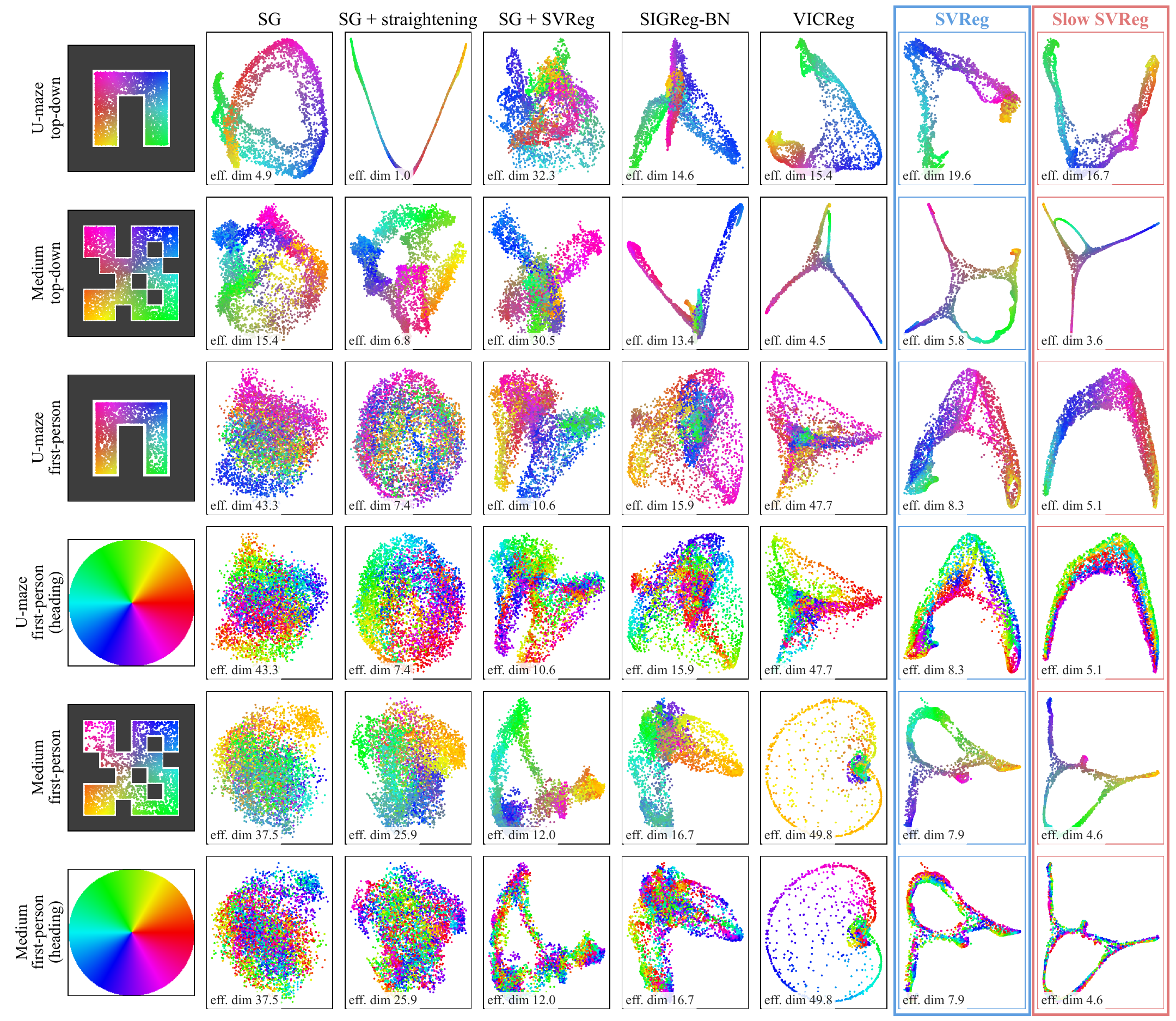}
\caption{\textbf{Latent PCA for top-down and first-person PointMaze.} SVReg and Slow SVReg preserve the map layout, and heading is also visible.
 }
\label{fig:maze_pca_all}
\end{figure}

\Cref{tab:maze_geometry} gives the path and distance measures of the maze latents in \Cref{fig:maze_pca} and \Cref{fig:maze_pca_all}.

\begin{table}[h]
\centering
\caption{\textbf{PointMaze latent geometry.} Settings for hard goals.}
\label{tab:maze_geometry}
\small
\begin{adjustbox}{max width=\linewidth}
\begin{tabular}{lcccccccc}
\toprule
 & \multicolumn{4}{c}{U-maze} & \multicolumn{4}{c}{Medium} \\
\cmidrule(lr){2-5}\cmidrule(lr){6-9}
method & plan (\%) & geodesic $\rho$ $\uparrow$ & local $\rho$ $\uparrow$ & jag.\ $\downarrow$ & plan (\%) & geodesic $\rho$ $\uparrow$ & local $\rho$ $\uparrow$ & jag.\ $\downarrow$ \\
\midrule
SG & 65.0\pmsd{18.7} & 0.68\pmsd{0.06} & 0.86\pmsd{0.03} & 5.03\pmsd{0.86} & 71.6\pmsd{8.7} & 0.63\pmsd{0.08} & 0.71\pmsd{0.12} & 4.48\pmsd{0.53} \\
SG + straightening & 88.0\pmsd{3.1} & \textbf{0.95\pmsd{0.01}} & 0.67\pmsd{0.02} & \textbf{1.02\pmsd{0.00}} & 72.8\pmsd{5.6} & 0.57\pmsd{0.08} & 0.82\pmsd{0.06} & 4.38\pmsd{0.52} \\
SG + SVReg & 95.8\pmsd{1.1} & 0.15\pmsd{0.01} & 0.78\pmsd{0.04} & 2.22\pmsd{0.46} & 62.7\pmsd{6.2} & 0.35\pmsd{0.09} & 0.82\pmsd{0.06} & 3.96\pmsd{0.49} \\
Two-stage (SG trunk + SVReg head)& 76.0\pmsd{20.6} & \textbf{0.95\pmsd{0.03}} & 0.85\pmsd{0.03} & 4.32\pmsd{1.08} & 83.0\pmsd{2.0} & 0.94\pmsd{0.01} & 0.71\pmsd{0.04} & 3.03\pmsd{0.48} \\
VICReg & 94.3\pmsd{2.3} & 0.63\pmsd{0.01} & 0.78\pmsd{0.03} & 1.90\pmsd{0.64} & 77.2\pmsd{2.3} & 0.92\pmsd{0.00} & 0.55\pmsd{0.03} & 2.29\pmsd{0.43} \\
SIGReg-BN & 82.2\pmsd{13.7} & 0.38\pmsd{0.07} & 0.77\pmsd{0.02} & 5.14\pmsd{0.43} & 78.2\pmsd{3.8} & 0.64\pmsd{0.02} & \textbf{0.89\pmsd{0.01}} & 1.65\pmsd{0.31} \\
\rowblue SVReg & 96.4\pmsd{1.3} & 0.85\pmsd{0.03} & 0.86\pmsd{0.02} & 1.51\pmsd{0.23} & 81.3\pmsd{2.3} & \textbf{0.95\pmsd{0.01}} & 0.80\pmsd{0.01} & 1.07\pmsd{0.04} \\
\rowblue Slow SVReg (ours) & 98.7\pmsd{0.7} & 0.89\pmsd{0.04} & \textbf{0.89\pmsd{0.01}} & 1.29\pmsd{0.13} & 84.7\pmsd{2.7} & 0.94\pmsd{0.00} & 0.70\pmsd{0.00} & \textbf{1.02\pmsd{0.01}} \\
\bottomrule
\end{tabular}
\end{adjustbox}
\end{table}

\paragraph{Reacher.}
A Reacher frame shows two joint angles. The shoulder turns fully and the elbow covers $\pm160^\circ$, so a faithful latent is a smooth surface with one loop for the shoulder. The elbow may close a second loop, because the two fully folded poses look alike. We pose the arm in the simulator, turn the shoulder through $360^\circ$ at a fixed elbow and project the frames onto each method's PCA. In \Cref{tab:reacher_geometry}, SVReg and SG + SVReg trace one smooth closed loop, while SG and two-stage trace jagged paths. As in the mazes, local smoothness tracks planning more closely than the global ranking of distances.

\begin{table}[h]
\centering
\caption{\textbf{Reacher latent geometry.}}
\label{tab:reacher_geometry}
\small
\begin{adjustbox}{max width=\linewidth}
\begin{tabular}{lcccc}
\toprule
method & plan (\%) & sweep jag.\ $\downarrow$ & dist.\ $\rho$ $\uparrow$ & 10-NN $\uparrow$ \\
\midrule
SG & 86.5\pmsd{1.6} & 3.48\pmsd{0.40} & 0.29\pmsd{0.02} & 0.66\pmsd{0.01} \\
SG + straightening & 86.6\pmsd{2.9} & 3.81\pmsd{0.45} & 0.18\pmsd{0.03} & 0.63\pmsd{0.02} \\
SG + SVReg & 96.3\pmsd{1.6} & \textbf{1.06\pmsd{0.02}} & 0.46\pmsd{0.08} & 0.70\pmsd{0.01} \\
Two-stage (SG trunk + SVReg head)& 92.7\pmsd{1.1} & 2.56\pmsd{0.22} & 0.47\pmsd{0.03} & 0.60\pmsd{0.03} \\
VICReg & 93.1\pmsd{1.7} & 1.37\pmsd{0.07} & 0.28\pmsd{0.01} & 0.68\pmsd{0.02} \\
SIGReg-BN & 95.8\pmsd{1.3} & 1.83\pmsd{0.14} & \textbf{0.62\pmsd{0.03}} & 0.65\pmsd{0.01} \\
\rowblue SVReg & 98.7\pmsd{0.9} & 1.10\pmsd{0.03} & 0.45\pmsd{0.06} & \textbf{0.72\pmsd{0.01}} \\
\rowblue Slow SVReg (ours) & 99.6\pmsd{0.4} & 1.12\pmsd{0.03} & 0.44\pmsd{0.02} & \textbf{0.72\pmsd{0.01}} \\
\bottomrule
\end{tabular}
\end{adjustbox}
\end{table}

\begin{figure}[p]
\centering
\includegraphics[width=\textwidth]{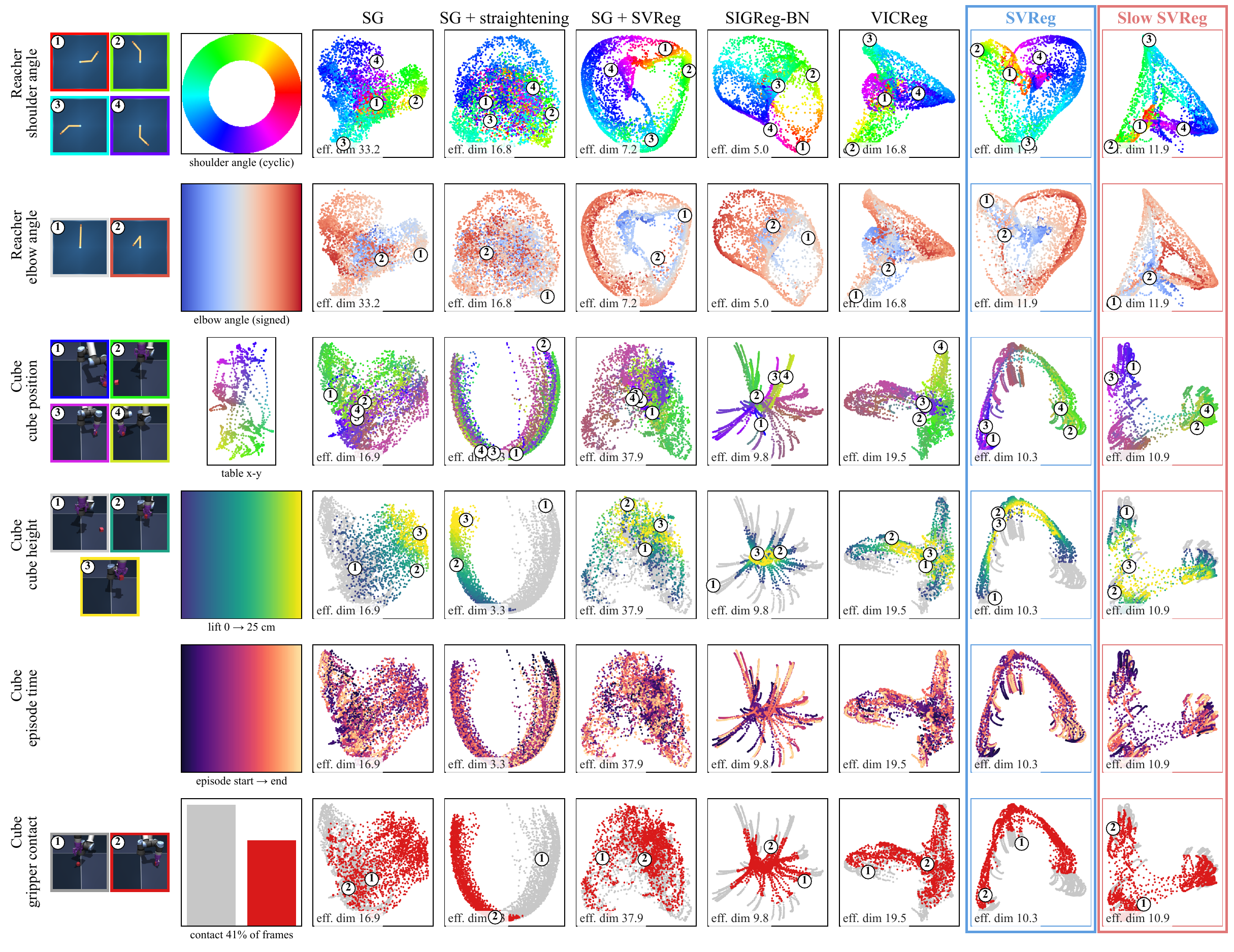}
\caption{\textbf{Latent PCA in Reacher and Cube.} Rows color held-out frames by joint angles (Reacher) and by cube position, lift, episode time and contact (Cube). Numbers mark the reference frames on the left. SVReg spreads the cube's table position over the leading components, and Slow SVReg also separates the lifted cube.}
\label{fig:manip_pca_all}
\end{figure}

\paragraph{Cube.}
The Cube state space has two sheets that meet at grasp and release. Without contact, arm and cube move independently, and the latent must keep the cube's position although nothing moves it. During a grasp, the cube moves with the gripper, so the latent should be a smooth function of the arm.
\Cref{tab:cube_geometry} gives ridge readouts on held-out episodes for the free cube, contact and geometry, and \Cref{fig:manip_pca_all} shows the latent PCA of all methods in Reacher and Cube. SIGReg-BN and VICReg keep cube height only in low-variance components. VICReg and SVReg are the only methods that fail to separate contact in some runs. SVReg merges the two sheets in three of six runs, which are also the runs that lose cube height.

\begin{table}[h]
\centering
\caption{\textbf{Cube latent geometry and content.}}
\label{tab:cube_geometry}
\small
\begin{adjustbox}{max width=\linewidth}
\begin{tabular}{lcccccc}
\toprule
method & plan (\%) & eff.\ dim & contact AUC $\uparrow$ & free cube $R^2$ $\uparrow$ & cost-to-goal $\rho$ $\uparrow$ & jag.\ $\downarrow$ \\
\midrule
SG & 80.4\pmsd{2.1} & 17.7\pmsd{0.6} & 0.99\pmsd{0.00} & 0.43\pmsd{0.11} & 0.38\pmsd{0.03} & 1.69\pmsd{0.03} \\
SG + straightening & 85.4\pmsd{1.0} & \phantom{0}2.9\pmsd{1.5} & 0.99\pmsd{0.01} & 0.51\pmsd{0.08} & 0.31\pmsd{0.09} & 1.19\pmsd{0.05} \\
SG + SVReg & 77.2\pmsd{1.3} & 35.4\pmsd{1.4} & 0.97\pmsd{0.02} & 0.45\pmsd{0.08} & 0.35\pmsd{0.02} & 2.19\pmsd{0.21} \\
Two-stage (SG trunk + SVReg head)& 86.8\pmsd{2.4} & \phantom{0}6.5\pmsd{0.2} & 0.99\pmsd{0.00} & \textbf{0.61\pmsd{0.03}} & \textbf{0.86\pmsd{0.02}} & 1.16\pmsd{0.01} \\
VICReg & 69.6\pmsd{5.8} & 19.8\pmsd{0.9} & 0.86\pmsd{0.08} & 0.58\pmsd{0.06} & 0.39\pmsd{0.03} & 1.63\pmsd{0.13} \\
SIGReg-BN & 87.4\pmsd{0.8} & \phantom{0}8.8\pmsd{0.6} & \textbf{1.00\pmsd{0.00}} & 0.59\pmsd{0.02} & 0.29\pmsd{0.04} & \textbf{1.11\pmsd{0.01}} \\
\rowblue SVReg & 76.9\pmsd{8.8} & 10.5\pmsd{1.1} & 0.80\pmsd{0.20} & 0.50\pmsd{0.05} & 0.76\pmsd{0.05} & 1.19\pmsd{0.11} \\
\rowblue Slow SVReg (ours) & 89.1\pmsd{1.6} & 12.0\pmsd{0.8} & 0.99\pmsd{0.01} & 0.57\pmsd{0.09} & 0.73\pmsd{0.05} & 1.12\pmsd{0.01} \\
\bottomrule
\end{tabular}
\end{adjustbox}
\end{table}

\begin{figure}[tbp]
\centering
\begin{minipage}[b]{0.41\linewidth}
\centering
\includegraphics[width=\linewidth]{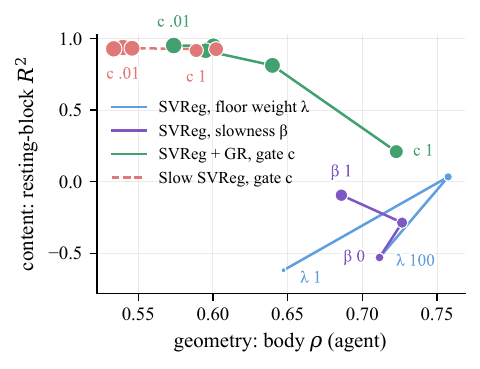}
\caption{\textbf{PushT content and geometry along single knobs.} $\rho$: rank correlation between latent and agent distances. Circle area: planning success.}
\label{fig:tradeoff_pusht}
\end{minipage}\hfill
\begin{minipage}[b]{0.55\linewidth}
\centering
\includegraphics[width=\linewidth]{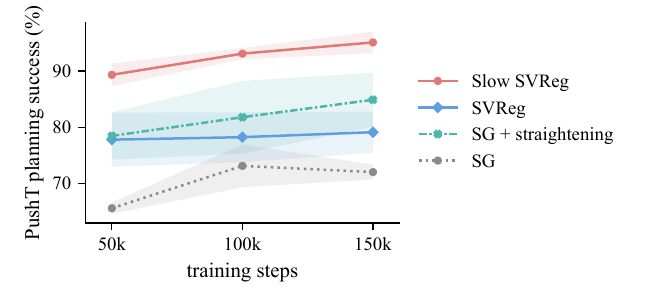}
\caption{\textbf{Slow SVReg improves with training steps on PushT.} Mean with $\pm$1 s.d. shaded. Each method trains on the full data at its \Cref{tab:main} setting and a constant learning rate for 150k steps. Each point branches off and decays the learning rate linearly to zero over its last 10\% of steps.}
\label{fig:stepscale}
\end{minipage}
\end{figure}

\paragraph{PushT.}
The PushT state is the agent position and the block pose.
In \Cref{tab:pusht_geometry}, SVReg's block readout does not transfer to new episodes in five of six runs, whether the block rests or is pushed, and its leading principal components carry the agent instead. \Cref{fig:tradeoff_pusht} traces content and geometry along single knobs on the full PushT dataset.

\begin{table}[h]
\centering
\caption{\textbf{PushT latent geometry and content.} SG (two-stage trunk): the frozen SG encoder of two-stage.}
\label{tab:pusht_geometry}
\small
\begin{adjustbox}{max width=\linewidth}
\begin{tabular}{lcccccc}
\toprule
method & plan (\%) & eff.\ dim & agent $R^2$ $\uparrow$ & free block $R^2$ $\uparrow$ & pushed block $R^2$ $\uparrow$ & cost-to-goal $\rho$ $\uparrow$ \\
\midrule
SG & 44.0\pmsd{19.3} & 13.4\pmsd{1.1} & 0.83\pmsd{0.04} & \phantom{$-$}0.20\pmsd{0.18} & \phantom{$-$}0.13\pmsd{0.20} & \phantom{$-$}0.55\pmsd{0.02}  \\
SG + straightening & 62.1\pmsd{7.7} & \phantom{0}4.8\pmsd{0.5} & 0.92\pmsd{0.01} & \phantom{$-$}0.55\pmsd{0.13} & \phantom{$-$}0.61\pmsd{0.09} & \phantom{$-$}0.68\pmsd{0.02}  \\
SG + SVReg & 76.9\pmsd{4.3} & 14.8\pmsd{1.2} & 0.89\pmsd{0.03} & \phantom{$-$}0.72\pmsd{0.07} & \phantom{$-$}0.80\pmsd{0.07} & \phantom{$-$}0.62\pmsd{0.04}  \\
SG (two-stage trunk, lr 1e-5) & 49.0\pmsd{6.1} & \phantom{0}8.8\pmsd{0.8} & 0.91\pmsd{0.02} & \phantom{$-$}0.72\pmsd{0.05} & \phantom{$-$}0.76\pmsd{0.08} & \phantom{$-$}0.80\pmsd{0.03}  \\
Two-stage (SG trunk + SVReg head)& 86.7\pmsd{3.0} & \phantom{0}7.1\pmsd{0.3} & 0.93\pmsd{0.01} & \textbf{\phantom{$-$}0.89\pmsd{0.04}} & \textbf{\phantom{$-$}0.91\pmsd{0.03}} & \textbf{\phantom{$-$}0.84\pmsd{0.03}}  \\
VICReg & 55.0\pmsd{3.6} & 36.9\pmsd{4.7} & 0.90\pmsd{0.05} & $-$0.57\pmsd{0.40} & $-$0.28\pmsd{0.38} & $-$0.01\pmsd{0.10}  \\
SIGReg-BN & 60.4\pmsd{2.6} & 24.7\pmsd{0.5} & \textbf{0.98\pmsd{0.00}} & $-$0.92\pmsd{0.23} & $-$0.32\pmsd{0.22} & \phantom{$-$}0.37\pmsd{0.05}  \\
\rowblue SVReg & 75.0\pmsd{8.6} & 13.0\pmsd{0.3} & 0.97\pmsd{0.04} & $-$0.27\pmsd{0.64} & \phantom{$-$}0.08\pmsd{0.41} & \phantom{$-$}0.67\pmsd{0.05}  \\
\rowblue Slow SVReg (ours) & 87.3\pmsd{3.8} & 10.8\pmsd{0.5} & 0.90\pmsd{0.06} & \phantom{$-$}0.79\pmsd{0.23} & \phantom{$-$}0.76\pmsd{0.26} & \phantom{$-$}0.71\pmsd{0.02}  \\
\bottomrule
\end{tabular}
\end{adjustbox}
\end{table}

\paragraph{Two-stage.}\label{app:twostage} Two-stage freezes an SG encoder and trains a fresh head, predictor and action encoder on its output with SVReg and without SG (\Cref{tab:arch}). \Cref{tab:twostage} compares it with SG and Slow SVReg on all tasks.

\begin{table}[!t]\centering\small\setlength{\tabcolsep}{4pt}
\caption{\textbf{Planning success (\%) of two-stage.}}
\label{tab:twostage}
\resizebox{\linewidth}{!}{%
\begin{tabular}{l*{8}{c}}
\toprule
 & \multicolumn{5}{c}{Agent only: navigation and reaching} & \multicolumn{3}{c}{Object manipulation} \\
\cmidrule(lr){2-6} \cmidrule(lr){7-9}
 & \multicolumn{2}{c}{U-maze (5k)} & \multicolumn{2}{c}{Medium (20k)} & Reacher (5k) & \multicolumn{2}{c}{Cube (80k)} & PushT (470k) \\
\cmidrule(lr){2-3} \cmidrule(lr){4-5} \cmidrule(lr){6-6} \cmidrule(lr){7-8} \cmidrule(lr){9-9}
Method & easy & hard & easy & hard & & all & $\geq$4\,cm & \\
\midrule
SG                            & 85.1\pmsd{17.0} & 65.0\pmsd{18.7} & 96.6\pmsd{3.1} & 71.6\pmsd{8.7} & 86.5\pmsd{1.6} & 80.4\pmsd{2.1} & 67.2\pmsd{3.6} & 44.0\pmsd{19.3} \\
Two-stage (SG trunk + SVReg head)         & 80.6\pmsd{21.7} & 76.0\pmsd{20.6} & 91.9\pmsd{0.8} & 83.0\pmsd{2.0} & 92.7\pmsd{1.1} & 86.8\pmsd{2.4} & 78.0\pmsd{3.6} & 86.7\pmsd{3.0} \\
\rowblue Slow SVReg (ours)             & {99.7}\pmsd{0.4} & {98.7}\pmsd{{0.7}} & {97.9}\pmsd{0.8} & {84.7}\pmsd{{2.7}} & {99.6}\pmsd{{0.4}} & {89.1}\pmsd{{1.6}} & {81.9}\pmsd{{2.5}} & {87.3}\pmsd{{3.8}} \\
\bottomrule
\end{tabular}}
\end{table}

\section{Content and Ablation Details}\label{app:ablation}
\paragraph{Content.} At the \Cref{tab:main} settings, Slow SVReg reads cube height ($R^2 \ge 0.98$) and separates contact in every run, and keeps the PushT block in five of six.

\paragraph{Ablation.} The gated residual alone fails, with Cube success below 80\% on all goals, in three of the six runs of \Cref{tab:ablation} ($\lambda=30$). Over 16 training seeds at $\lambda=10$, the gated residual alone fails in four and Slow SVReg in none. On PushT, the resting-block $R^2$ is 0.12, 0.59 and 0.79 with slow channels, the gated residual and both.

\paragraph{Early phase.} We rerun four seeds of each variant at $\lambda=10$ with intermediate checkpoints and measure jitter as the variance-weighted share of latent power at periods under five model steps on held-out trajectories. At step 5000, the gated residual alone is jittery in three seeds and reads height less well ($R^2$ 0.54--0.79), while Slow SVReg is already smooth ($R^2$ 0.99). The two failing seeds recover by 100k steps (89\% and 87\%). The slow channels hold almost none of the height readout's weight, and slowness on the whole latent without a slow block also removes all failures over the 16 seeds.

\paragraph{Controls.} The element-wise floor plans at about 50\% on Cube and reads cube height poorly ($R^2 < 0.5$) at every $\lambda\in\{1,10,100\}$. SG with the gated residual reads no cube height in any run.

\section{Switching Objectives}\label{app:switch}
\begin{figure}[!tbp]
\centering
\begin{minipage}[c]{0.46\linewidth}
\centering
\includegraphics[width=\linewidth]{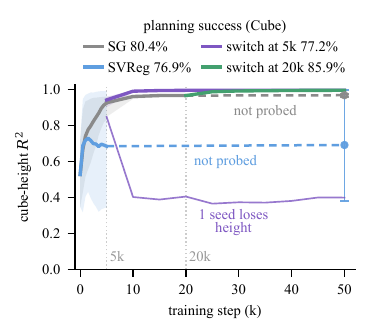}
\end{minipage}\hfill
\begin{minipage}[c]{0.5\linewidth}
\caption{\textbf{Cube-height readout after switching from SG to SVReg.} Mean and range over runs. Legend: planning success. One 5k-switch run is drawn apart. Dashed segments were not probed.}
\label{fig:switch}
\end{minipage}
\end{figure}
Switching from SG to SVReg tests whether acquired content persists. In \Cref{fig:switch}, switching at 20k steps keeps cube height in all runs, while switching at 5k loses it in one.

\section{Data Scaling}\label{app:scaling}
Each scaling run repeats a \Cref{tab:main} run with less training data and keeps its setting, validation data, number of training steps and planning protocol. Each dataset is a prefix of the original collection, a single exploration trajectory in the mazes and Reacher and a set of expert episodes on Cube. The mazes are scored on hard goals and Cube on goals that move the cube by at least 4\,cm, as in \Cref{tab:main}. Each point uses three training runs, and the full-data points use six.

\begin{figure}[h]
\centering
\includegraphics[width=\linewidth]{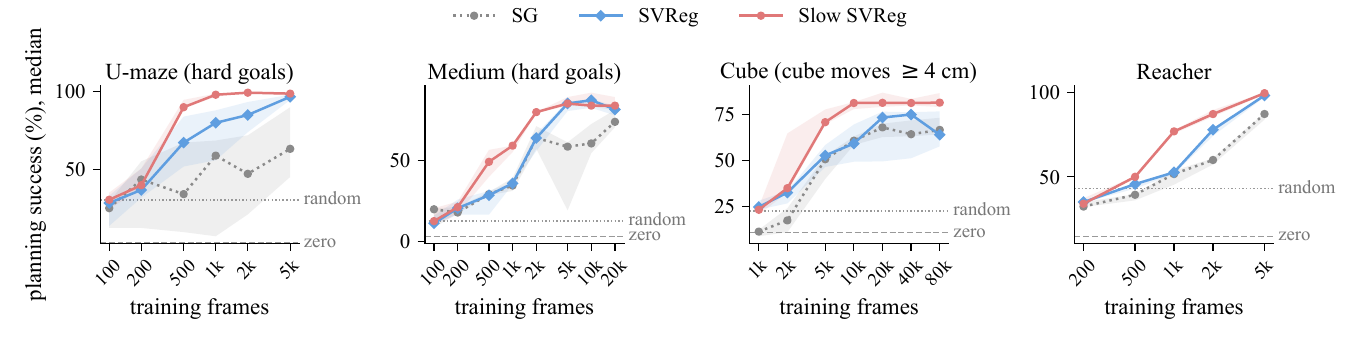}
\caption{\textbf{Planning success against dataset size.} Median over training seeds with the range shaded. Each panel ends at the \Cref{tab:main} dataset size.}
\label{fig:scaling}
\end{figure}

\section{City Walks}\label{app:city}
\paragraph{Data.} The random walker stays within 1.2\,m of the capture path, where renders stay sharp, and moves into nearby geometry are refused using the rendered depth. The main scene, 13Skv, has 100 training walks of 500 steps, 10 validation walks and a set of held-out walks. A disc of radius one seventh of the walkable extent is held out of all training walks and is scored as a separate set.

\paragraph{Models.} The encoder, predictor and objectives follow the maze setup. SIGReg-BN uses per-frame statistics in City Walks, as in LeWM. The untrained control has random weights and BatchNorm statistics recalibrated on the training walks (three initializations). Predicting the mean position gives a 28.0\,m error.

\begin{figure}[t]
\centering
\sbox0{\includegraphics[width=\linewidth]{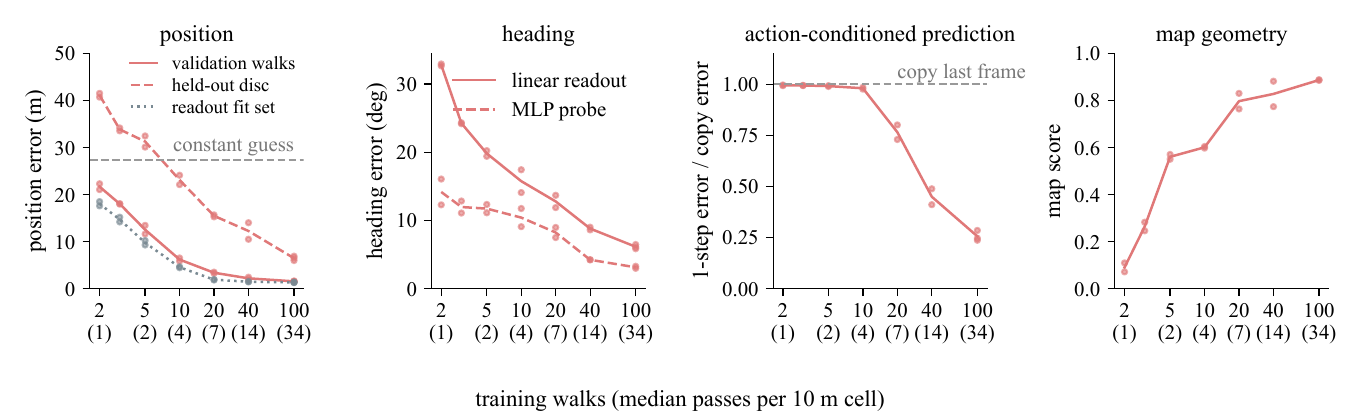}}\usebox0\plab{0.006}{0.92}{a}\plab{0.257}{0.92}{b}\plab{0.497}{0.92}{c}\plab{0.756}{0.92}{d}
\caption{\textbf{City-map measures against the number of training walks.} Slow SVReg on scene 13Skv. Dots: runs. (a)~Position error. (b)~Heading error. (c)~One-step prediction error relative to copying the last frame. (d)~Map score.}
\label{fig:walk_sweep}
\end{figure}

\begin{table}[t]\centering\small\setlength{\tabcolsep}{4pt}
\caption{\textbf{Latent maps on ten further Wanderland scenes.} Separate models per scene at the scene-13Skv setting, mean of two runs. Bold: best per scene.}
\label{tab:city_multi}
\begin{adjustbox}{max width=\linewidth}
\begin{tabular}{l*{6}{c}@{\hspace{10pt}}*{6}{c}}
\toprule
 & \multicolumn{6}{c}{position (m) $\downarrow$} & \multicolumn{6}{c}{map score $\uparrow$} \\
\cmidrule(r{6pt}){2-7} \cmidrule{8-13}
Scene & SG & SG + str. & VICReg & SIGReg-BN & SVReg & Slow SVReg & SG & SG + str. & VICReg & SIGReg-BN & SVReg & Slow SVReg \\
\midrule
\multicolumn{13}{l}{\emph{Outdoor}} \\
11H3 & 2.6 & 5.5 & 6.3 & 1.4 & 2.7 & \textbf{1.1} & 0.14 & 0.10 & 0.06 & 0.10 & 0.74 & \textbf{0.84} \\
1HUMP & 5.7 & 3.1 & 7.8 & \textbf{0.7} & 2.1 & 1.0 & 0.33 & 0.32 & 0.09 & 0.37 & 0.72 & \textbf{0.79} \\
18qtBG3q & 3.4 & 4.5 & 5.9 & \textbf{1.0} & 3.2 & 1.3 & 0.02 & 0.19 & 0.11 & 0.30 & 0.63 & \textbf{0.72} \\
1IjqGTz0 & 8.1 & 8.2 & 8.5 & \textbf{7.9} & 10.0 & 10.0 & 0.18 & 0.41 & 0.09 & 0.20 & 0.56 & \textbf{0.59} \\
1L58FmPh & 1.9 & 1.3 & 3.2 & \textbf{0.9} & 2.0 & \textbf{0.9} & 0.08 & 0.39 & 0.15 & 0.26 & \textbf{0.58} & 0.53 \\
1AlI8jGb & 5.8 & 9.1 & 4.5 & 1.1 & 3.7 & \textbf{1.0} & 0.01 & 0.00 & 0.09 & 0.23 & 0.67 & \textbf{0.72} \\
\multicolumn{13}{l}{\emph{Indoor}} \\
1\_nV & 8.6 & 5.3 & 5.7 & \textbf{2.7} & 5.7 & 3.5 & 0.03 & 0.14 & 0.15 & 0.05 & 0.68 & \textbf{0.77} \\
12Qy & 8.7 & 6.9 & 5.6 & \textbf{3.5} & 7.1 & 6.6 & 0.04 & 0.21 & 0.16 & 0.20 & 0.21 & \textbf{0.32} \\
1yPum & 5.0 & 4.6 & 3.8 & \textbf{1.3} & 2.6 & 3.7 & 0.05 & 0.06 & 0.05 & 0.04 & \textbf{0.40} & 0.23 \\
1qbLFA84 & 3.8 & 3.5 & 2.1 & \textbf{1.0} & 1.9 & 2.3 & 0.15 & 0.24 & 0.36 & 0.39 & \textbf{0.79} & \textbf{0.79} \\
\midrule
Mean & 5.4 & 5.2 & 5.3 & \textbf{2.1} & 4.1 & 3.1 & 0.10 & 0.21 & 0.13 & 0.21 & 0.60 & \textbf{0.63} \\
Median & 5.4 & 4.9 & 5.7 & \textbf{1.2} & 2.9 & 1.8 & 0.06 & 0.20 & 0.10 & 0.21 & 0.65 & \textbf{0.72} \\
\bottomrule
\end{tabular}
\end{adjustbox}
\end{table}

\paragraph{Selection.} We select each method's learning rate and regularization weight by the lowest validation position error on one training seed. \Cref{tab:city} averages the selected setting over three further runs, and \Cref{fig:city_map} shows the best of them. The ten further scenes use two runs each.

\paragraph{Other variants.} With pooled statistics, SIGReg-BN localizes less accurately and forms no map on any of the eleven scenes. On 13Skv, two-stage (\Cref{app:twostage}, one setting) maps nearly as clearly as Slow SVReg but reads position less accurately, and SG + straightening reads position in two of three runs.

\paragraph{Metrics.} Position and heading come from ridge readouts fit on 6{,}000 training-walk frames and scored on the validation walks and the held-out disc. The map score is the Procrustes $R^2$ between the top two principal components of the validation latents and the true positions: 1 for a rotated and scaled copy of the walked area, 0 for no layout. For cross-heading recall (\Cref{sec:city}), chance is 4\% and the untrained encoder reaches 5\%. On scene 13Skv, Slow SVReg, SIGReg-BN and SVReg reach 75--89\%, VICReg and two-stage 26--36\%, and SG + straightening and SG about 10\%. The ten further scenes give a similar ranking.

\end{document}